# Preserving privacy in domain transfer of medical AI models comes at no performance costs: The integral role of differential privacy

Soroosh Tayebi Arasteh (1), Mahshad Lotfinia (1,2), Teresa Nolte (1), Marwin Saehn (1), Peter Isfort (1), Christiane Kuhl (1), Sven Nebelung (1), Georgios Kaissis* (3,4,5), Daniel Truhn* (1)

(1) Department of Diagnostic and Interventional Radiology, University Hospital RWTH Aachen, Aachen, Germany.
(2) Institute of Heat and Mass Transfer, RWTH Aachen University, Aachen, Germany.
(3) Institute of Diagnostic and Interventional Radiology, Technical University of Munich, Munich, Germany.
(4) Artificial Intelligence in Healthcare and Medicine, Technical University of Munich, Munich, Germany.
(5) Institute for Machine Learning in Biomedical Imaging, Helmholtz-Zentrum Munich, Neuherberg, Germany.

* G.K. and D.T. are co-senior authors.

## Abstract

Developing robust and effective artificial intelligence (AI) models in medicine requires access to large amounts of patient data. The use of AI models solely trained on large multi-institutional datasets can help with this, yet the imperative to ensure data privacy remains, particularly as membership inference risks breaching patient confidentiality. As a proposed remedy, we advocate for the integration of differential privacy (DP). We specifically investigate the performance of models trained with DP as compared to models trained without DP on data from institutions that the model had not seen during its training (i.e., external validation) - the situation that is reflective of the clinical use of AI models. By leveraging more than 590,000 chest radiographs from five institutions, we evaluated the efficacy of DP-enhanced domain transfer (DP-DT) in diagnosing cardiomegaly, pleural effusion, pneumonia, atelectasis, and in identifying healthy subjects. We juxtaposed DP-DT with non-DP-DT and examined diagnostic accuracy and demographic fairness using the area under the receiver operating characteristic curve (AUC) as the main metric, as well as accuracy, sensitivity, and specificity. Our results show that DP-DT, even with exceptionally high privacy levels ($\varepsilon \approx 1$), performs comparably to non-DP-DT ($P > 0.119$ across all domains). Furthermore, DP-DT led to marginal AUC differences - less than 1% - for nearly all subgroups, relative to non-DP-DT. Despite consistent evidence suggesting that DP models induce significant performance degradation for on-domain applications, we show that off-domain performance is almost not affected. Therefore, we ardently advocate for the adoption of DP in training diagnostic medical AI models, given its minimal impact on performance.

# Introduction

As the amount of patient data is ever-increasing, employing artificial intelligence (AI) models becomes more appealing since performance and capabilities of these models will presumably rise. The prevailing strategy to increase the amount of training data is collaboration between institutions. If models can be trained on big multi-institutional datasets, they perform better on new external datasets[1–4]. We refer to this paradigm as "domain transfer" performance, i.e., the process of testing the trained models on samples from a different distribution.

However, strict data sharing policies rightfully prevent unconditional access to patient data by external institutions for the training of deep learning models. Therefore, privacy-preserving methods to train AI models are needed. Federated learning[5–9] offers a potential solution since it does not require the data provider to transfer the data. However, it has consistently been shown that federated learning is not truly privacy-preserving, as network parameters and gradients are vulnerable to breach of information through membership inference and reconstruction attacks[10–12] (**Figure 1**).

A key-method for guaranteeing patient privacy is differential privacy[13] (DP), which has gained a lot of attention in the AI community recently[11,14]. DP is a concept of formally quantifying privacy guarantees (e.g., by using parameters ε / δ) and obtaining insights from sensitive datasets while protecting individual data points within them[11,13]. Within the setting of deep learning, differentially-private stochastic gradient descent[15] is a training paradigm that adds calibrated noise to gradients during the training process and limits the amount of information that gradient updates carry. However, higher privacy levels may come with a privacy-performance trade-off and a privacy-fairness trade-off [16,17]. Previous research[11] empirically examined DP training of diagnostic AI models using a large chest radiograph cohort for single-institutional application and observed a small yet significant drop in diagnostic performance, while maintaining guaranteed privacy.

In this study, we focused on domain transferability of models that have been trained with DP. We assessed whether training with DP impacted domain transfer performance. It is important to note that we examined only the transfer of AI models that are guaranteed to be privacy-preserving to external partners. Consequently, we performed a large-scale analysis of DP-enhanced domain transfer (DP-DT) using a total of n=590,000 radiographs from five institutions, covering a variety of different imaging settings.

To the best of our knowledge, this is the first analysis of DP in external domains for medical AI models. In particular, we compared the performance of diagnostic AI models in external domains trained with and without DP. We hypothesized that DP maintains - and potentially even increases – cross-institutional diagnostic performance due to the regularization effect of DP and since the formal guarantees of DP do not impact generalization if sufficient data is available. We also performed a detailed investigation to determine if DP leads to a reduction in performance for underrepresented groups, as this is a concern that has been hypothesized in the literature[18] and is particularly accentuated in medical AI models due to the potential impact on patients' health[19].

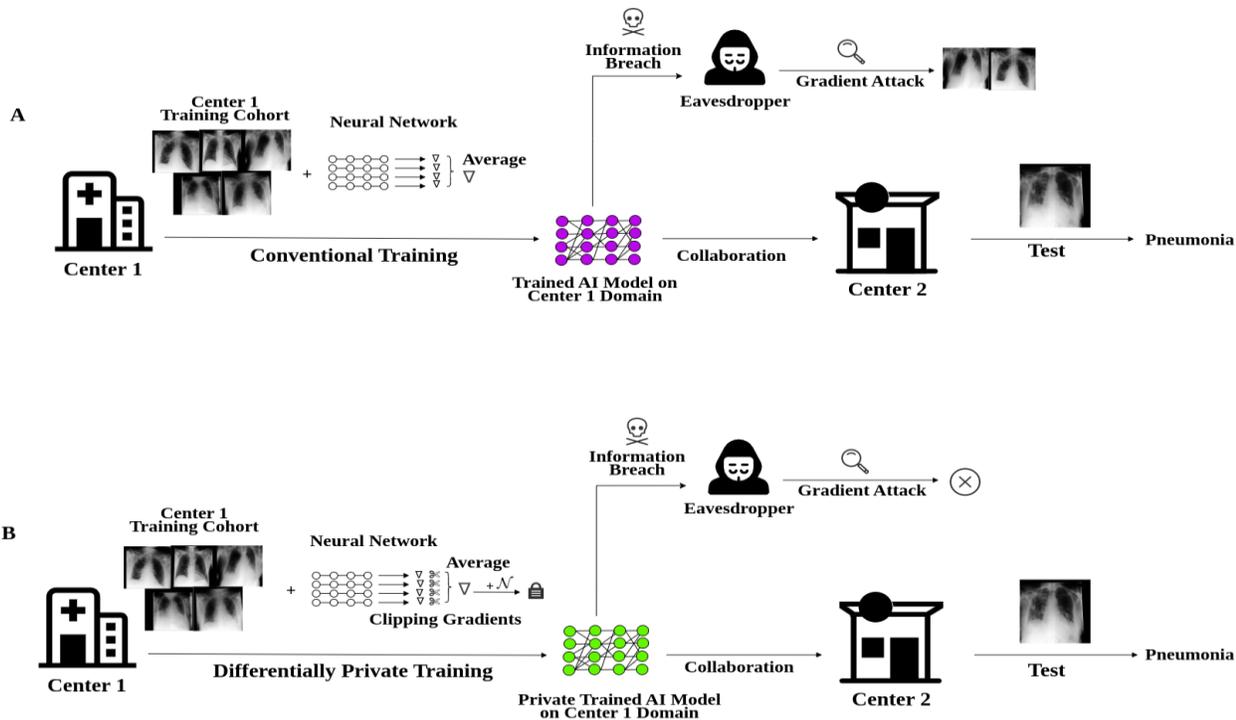

**Figure 1**: **The problem of conventional collaboration between institutions for sharing AI models. (A)** Center 1 conventionally trains an AI model for supine chest radiographs using its own data of patients in intensive care. It aims at sharing the AI model with another hospital for diagnosis of radiographs taken in upright position. However, an eavesdropper gets access to this model because of information breach and reconstructs the images of center 1 that were used for training that model. **(B)** Diagram shows a similar scenario, with the difference being that Center 1 shares a differentially-private-trained model of supine chest radiographs from patients in intensive care with Center 2. In this case, while Center 2 can perform diagnosis using this AI model on its upright radiographs, neither center 2 nor an eavesdropper can reconstruct the original training images of Center 1 due to the privacy-preserving nature of the model.

# Results

## DP-DT and Non-DP-DT Achieve Similar Diagnostic Performance

**Figure 2** shows the diagnostic performance of DP-networks across varying privacy budgets in different external domains, averaged over all labels, including cardiomegaly, pleural effusion, pneumonia, atelectasis, and healthy. The results demonstrate no evidence of differences in performance between the different $\varepsilon$ budgets, ranging from $\varepsilon \approx 1$ to $\varepsilon = \infty$ (non-DP). **Table 2** presents a more detailed comparison, showing the differences in AUC between two scenarios: i) DP-DT with a comparatively high privacy level of $\varepsilon \approx 1$ (VinDr-CXR: $\varepsilon = 1.17$, ChestX-ray14: $\varepsilon = 1.01$, CheXpert: $\varepsilon = 0.98$, UKA-CXR: $\varepsilon = 0.98$, and PadChest: $\varepsilon = 0.72$) and $\delta = 6 \times 10^{-6}$ and ii) non-DP-DT ($\varepsilon = \infty$), as compared to conventional single-institutional training without privacy measures. **Table S1** reports further evaluation metrics in terms of accuracy, sensitivity, and specificity, comparing DP-DT with non-DP-DT. For a better comparison, we measured the decrease in AUC of cross-institutional performance as compared with single-institutional performance for both setups. On average, these values were consistent between DP-DT and non-DP-DT and no evidence of differences were found (VinDr-CXR:

0.07 vs. 0.07; P=0.96, ChestX-ray14: 0.07 vs. 0.06; P=0.12, CheXpert: 0.07 vs. 0.07; P=0.18, UKA-CXR: 0.18 vs. 0.18; P=0.90, and PadChest: 0.07 vs. 0.07; P=0.35).

**Figures 3** and **S7–S9** showcase the diagnostic performance of DP networks for individual diseases for different ε values in various external domains. **Table S2** compares the cross-institutional performance of networks between DP-DT with $\varepsilon \approx 1$ and non-DP-DT for individual labels. We observed no evidence of differences between DP-DT and non-DP-DT for all individual labels (pleural effusion: 0.85 vs. 0.86; P=0.55, pneumonia: 0.74 vs. 0.74; P=0.82, atelectasis: 0.67 vs. 0.67; P=0.96, and healthy: 0.77 vs. 0.77; P=0.70), except for cardiomegaly (0.81 vs. 0.83; P=0.01).

**Table 1**: **Characteristics of the datasets used in the study.** Statistics of the datasets used, including VinDr-CXR[20], ChestX-ray14[21], CheXpert[22], UKA-CXR[9,11,23], and PadChest[24]. $\delta = 0.000006$ was selected for all datasets. The datasets included only anteroposterior and posteroanterior chest radiographs, with the percentages of total radiographs provided. Note that some datasets may include multiple radiographs per patient. N/A=not available, NLP=natural language processing.

|  | VinDr-CXR | ChestX-ray14 | CheXpert | UKA-CXR | PadChest |
|---|---|---|---|---|---|
| Number of Radiographs Total (training set/test set) [n] | 18,000 (15,000 / 3,000) | 112,120 (86,524 / 25,596) | 157,878 (128,356 / 29,320) | 193,361 (153,537 / 39,824) | 110,525 (88,480 / 22,045) |
| Number of Patients [n] | N/A | 30,805 | 65,240 | 54,176 | 67,213 |
| PATIENT AGE [years] Median Mean ± Standard Deviation Range (minimum, maximum) | 42 54 ± 18 (2, 91) | 49 47 ± 17 (1, 96) | 61 60 ± 18 (18, 91) | 68 66 ± 15 (1, 111) | 63 59 ± 20 (1, 105) |
| PATIENT SEX Females / Males [%] (training set, test set) | 47.8 / 52.2 44.1 / 55.9 | 42.4 / 57.6 41.9 / 58.1 | 41.4 / 58.6 39.0 / 61.0 | 34.4 / 65.6 36.3 / 63.7 | 50.0 / 50.0 48.2 / 51.8 |
| PROJECTIONS [%] anteroposterior posteroanterior | 0.0 100.0 | 40.0 60.0 | 84.5 15.5 | 100.0 0.0 | 17.1 82.9 |
| Country | Vietnam | USA | USA | Germany | Spain |
| Contributing Hospitals [n] | 2 | 1 | 1 | 1 | 1 |
| Labeling Method | Expert radiologists | NLP | NLP | Expert radiologists | Partially expert radiologists, partially NLP |
| Radiographs with cardiomegaly [%] | 11.8 | 2.5 | 12.6 | 46.7 | 8.9 |
| Radiographs with pleural effusion [%] | 4.1 | 11.9 | 41.3 | 12.6 | 6.3 |
| Radiographs with pneumonia [%] | 4.0 | 1.3 | 2.5 | 14.0 | 4.7 |
| Radiographs with atelectasis [%] | 0.8 | 10.3 | 16.7 | 14.0 | 5.6 |
| Radiographs with healthy label [%] | 70.3 | 53.8 | 10.8 | 38.5 | 32.7 |

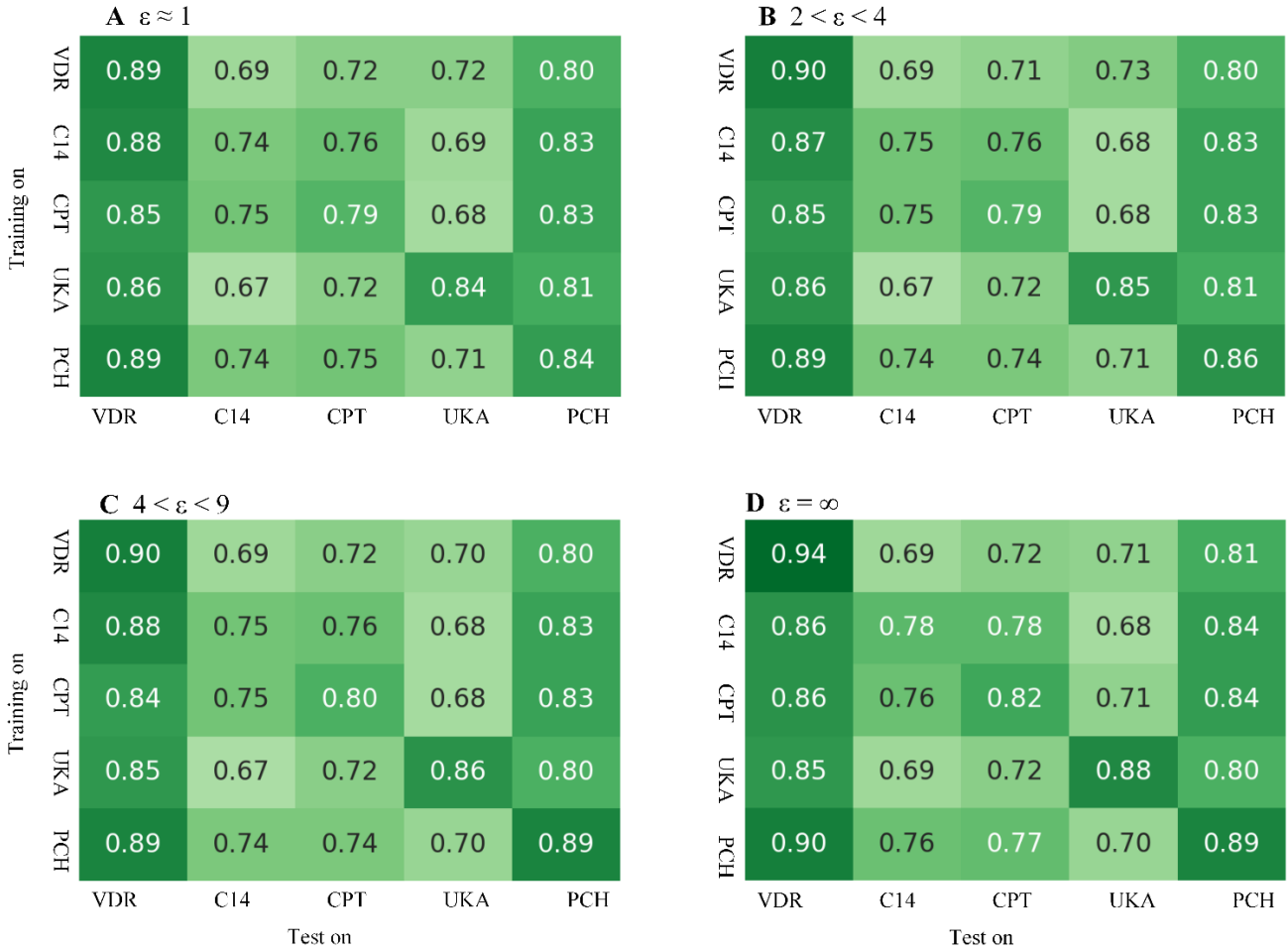

**Figure 2**: **Results of transferring DP models with different ε values to different domains.** The values correspond to average area under the receiver operating characteristic curve (AUC) results over all labels using networks with **(A)** ε≈1, **(B)** 2<ε<4, **(C)** 4<ε<9, and **(D)** ε=∞ (non-DP). Each row corresponds to a training domain, and each column corresponds to a test domain. The privacy budgets of the DP networks corresponding to each dataset for **(A)-(C)** are as follows: VinDr-CXR (VDR): ε=1.17, 3.24, and 4.29; ChestX-ray14 (C14): ε=1.01, 3.37, and 7.83; CheXpert (CPT): ε=0.98, 3.30, and 6.48; UKA-CXR (UKA): ε=0.98, 3.46, and 8.81; and PadChest (PCH): ε=0.72, 3.58, and 7.41.

**Table 2: Comparison of DP-Domain Transfer and non-DP-Domain Transfer Cross-Institutional Performance.** AUC values for each training-testing combination and their respective difference to the single-institutional testing AUC. AUC is averaged over all one-vs-all classifications (cardiomegaly, pleural effusion, pneumonia, atelectasis, and healthy). Note that there is no evidence of a difference between the DP-trained and the non-DP-trained networks. The privacy budgets of the DP networks were as follows: VinDr-CXR: ε=1.17, ChestX-ray14: ε=1.01, CheXpert: ε=0.98, UKA-CXR: ε=0.98, and PadChest: ε=0.72, with δ=0.000006 for all datasets. SI=single-institutional.

| | Test on: VinDr-CXR | | Test on: ChestX-ray14 | | Test on: CheXpert | | Test on: UKA-CXR | | Test on: PadChest | |
|---|---|---|---|---|---|---|---|---|---|---|
| | AUC [95% CI] | Diff | AUC [95% CI] | Diff | AUC [95% CI] | Diff | AUC [95% CI] | Diff | AUC [95% CI] | Diff |
| Baseline | 0.94 [0.93, 0.94] | 0.00 | 0.78 [0.77, 0.78] | 0.00 | 0.82 [0.81, 0.82] | 0.00 | 0.88 [0.88, 0.88] | 0.00 | 0.89 [0.89, 0.89] | 0.00 |
| ε≈1 (DP) | | | | | | | | | | |
| Training on: VinDr-CXR | SI | | 0.69 [0.69, 0.70] | 0.09 | 0.72 [0.71, 0.72] | 0.10 | 0.72 [0.72, 0.72] | 0.16 | 0.80 [0.80, 0.81] | 0.09 |
| Training on: ChestX-ray14 | 0.88 [0.87, 0.89] | 0.06 | SI | | 0.76 [0.75, 0.76] | 0.06 | 0.69 [0.69, 0.70] | 0.19 | 0.83 [0.82, 0.83] | 0.06 |
| Training on: CheXpert | 0.85 [0.83, 0.86] | 0.09 | 0.75 [0.75, 0.76] | 0.03 | SI | | 0.68 [0.68, 0.69] | 0.20 | 0.83 [0.82, 0.83] | 0.06 |
| Training on: UKA-CXR | 0.86 [0.85, 0.88] | 0.08 | 0.67 [0.67, 0.68] | 0.11 | 0.72 [0.72, 0.73] | 0.10 | SI | | 0.81 [0.80, 0.81] | 0.08 |
| Training on: PadChest | 0.89 [0.87, 0.90] | 0.05 | 0.74 [0.73, 0.74] | 0.04 | 0.75 [0.75, 0.76] | 0.07 | 0.71 [0.71, 0.72] | 0.17 | SI | |
| *Average* | *0.87 [0.85, 0.88]* | *0.07* | *0.71 [0.71, 0.72]* | *0.07* | *0.74 [0.73, 0.74]* | *0.07* | *0.70 [0.70, 0.71]* | *0.18* | *0.82 [0.81, 0.82]* | *0.07* |
| ε=∞ (Non-DP) | | | | | | | | | | |
| Training on: VinDr-CXR | SI | | 0.69 [0.68, 0.69] | 0.09 | 0.72 [0.71, 0.72] | 0.10 | 0.71 [0.70, 0.71] | 0.17 | 0.81 [0.81, 0.82] | 0.08 |
| Training on: ChestX-ray14 | 0.86 [0.84, 0.87] | 0.08 | SI | | 0.78 [0.77, 0.78] | 0.04 | 0.68 [0.68, 0.69] | 0.20 | 0.84 [0.84, 0.85] | 0.05 |
| Training on: CheXpert | 0.86 [0.85, 0.88] | 0.08 | 0.76 [0.76, 0.77] | 0.02 | SI | | 0.71 [0.70, 0.71] | 0.17 | 0.84 [0.83, 0.84] | 0.05 |
| Training on: UKA-CXR | 0.85 [0.83, 0.87] | 0.09 | 0.69 [0.68, 0.69] | 0.09 | 0.72 [0.72, 0.73] | 0.10 | SI | | 0.80 [0.80, 0.81] | 0.09 |
| Training on: PadChest | 0.90 [0.89, 0.91] | 0.04 | 0.76 [0.76, 0.77] | 0.02 | 0.77 [0.76, 0.77] | 0.05 | 0.70 [0.70, 0.70] | 0.18 | SI | |
| *Average* | *0.87 [0.85, 0.88]* | *0.07* | *0.73 [0.72, 0.73]* | *0.06* | *0.75 [0.74, 0.75]* | *0.07* | *0.70 [0.69, 0.70]* | *0.18* | *0.82 [0.82, 0.83]* | *0.07* |
| P-value | 0.96 | | 0.12 | | 0.18 | | 0.90 | | 0.35 | |

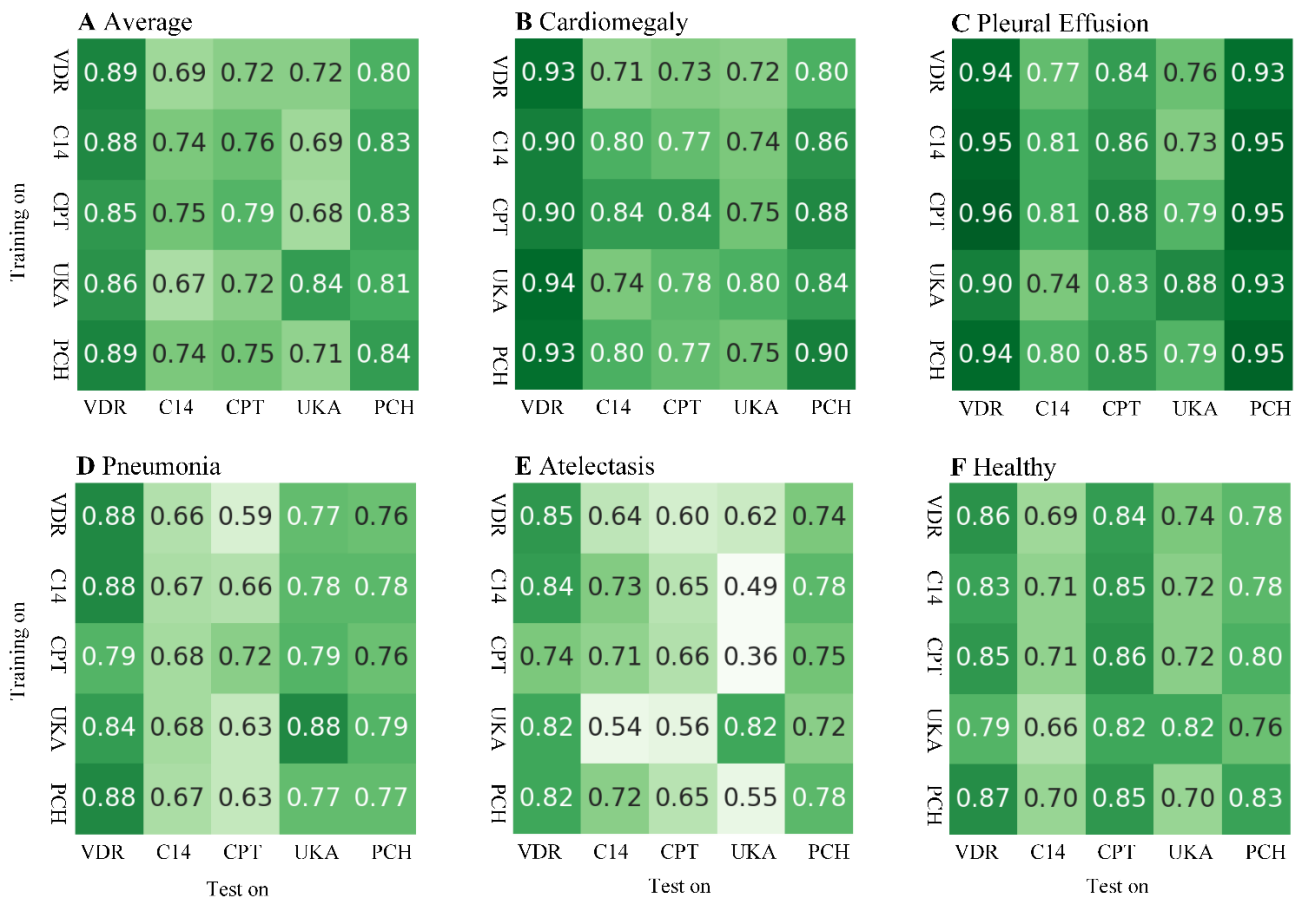

**Figure 3**: **Results of transferring DP models with ε≈1 to different domains for individual labels.** The area under the receiver operating characteristic curve (AUC) values correspond to **(A)** average over all labels, **(B)** cardiomegaly, **(C)** pleural effusion, **(D)** pneumonia, **(E)** atelectasis, and **(F)** healthy. Each row corresponds to a training domain and each column corresponds to a test domain. The privacy budgets of the DP networks corresponding to each dataset are as follows: VinDr-CXR (VDR): ε=1.17, ChestX-ray14 (C14): ε=1.01, CheXpert (CPT): ε=0.98, UKA-CXR (UKA): ε=0.98, and PadChest (PCH): ε=0.72, with δ=0.000006 for all datasets.

## DP-DT Has No Effect on Sex-Based Fairness

There has been concern that the application of DP can lead to decreased performance in groups that are underrepresented in the dataset[18].

To test this, we performed a sub-analysis in male and female patients. **Figure 4** and **Table 3** demonstrate that on average, DP-DT resulted in less than 1% AUC difference as compared with non-DP-DT for female and male subgroups in all datasets and no evidence of differences were found (VinDr-CXR: P=0.46 for female and P=0.22 for male; ChestX-ray14: P=0.45 for female and P=0.37 for male; CheXpert: P=0.39 for female and P=0.29 for male; UKA-CXR: P=0.40 for female and P=0.33 for male; PadChest: P=0.43 for female and P=0.22 for male). A more detailed analysis on each sex subgroup, including further evaluation metrics in terms of accuracy, sensitivity, and specificity, comparing DP-DT with non-DP-DT is reported in **Tables S3** and **S4**, and statistical parity difference values for individual sex subgroups are reported in **Table S5**.

**Table 3: Comparison of DP-Domain Transfer and non-DP-Domain Transfer Cross-Institutional Performance by Sex Subgroups.** The values in each column correspond to the average of AUC differences from the non-DP single institutional case for that test dataset (baseline) and are averaged over all labels including cardiomegaly, pleural effusion, pneumonia, atelectasis, and healthy. The privacy budgets of the DP networks corresponding to each dataset are as follows: VinDr-CXR: $\varepsilon=1.17$, ChestX-ray14: $\varepsilon=1.01$, CheXpert: $\varepsilon=0.98$, UKA-CXR: $\varepsilon=0.98$, and PadChest: $\varepsilon=0.72$, with $\delta=0.000006$ for all datasets. P-values are calculated between DP and non-DP methods. DP=differential privacy, F=female, M=male, SI=single-institutional.

|  | Test on: VinDr-CXR | | Test on: ChestX-ray14 | | Test on: CheXpert | | Test on: UKA-CXR | | Test on: PadChest | |
|---|---|---|---|---|---|---|---|---|---|---|
|  | F | M | F | M | F | M | F | M | F | M |
| $\varepsilon\approx 1$ (DP) | | | | | | | | | | |
| Training on: VinDr-CXR | SI | | 0.09 | 0.09 | 0.09 | 0.11 | 0.17 | 0.16 | 0.09 | 0.09 |
| Training on: ChestX-ray14 | 0.10 | 0.07 | SI | | 0.06 | 0.07 | 0.19 | 0.19 | 0.07 | 0.06 |
| Training on: CheXpert | 0.13 | 0.12 | 0.03 | 0.03 | SI | | 0.20 | 0.20 | 0.06 | 0.06 |
| Training on: UKA-CXR | 0.10 | 0.11 | 0.11 | 0.11 | 0.09 | 0.10 | SI | | 0.08 | 0.09 |
| Training on: PadChest | 0.07 | 0.09 | 0.04 | 0.05 | 0.06 | 0.07 | 0.17 | 0.17 | SI | |
| *Average* | *0.10* | *0.10* | *0.07* | *0.07* | *0.08* | *0.09* | *0.18* | *0.18* | *0.08* | *0.08* |
| $\varepsilon=\infty$ (Non-DP) | | | | | | | | | | |
| Training on: VinDr-CXR | SI | | 0.09 | 0.10 | 0.10 | 0.10 | 0.19 | 0.17 | 0.08 | 0.08 |
| Training on: ChestX-ray14 | 0.12 | 0.12 | SI | | 0.04 | 0.05 | 0.19 | 0.20 | 0.05 | 0.05 |
| Training on: CheXpert | 0.13 | 0.08 | 0.02 | 0.02 | SI | | 0.17 | 0.18 | 0.06 | 0.05 |
| Training on: UKA-CXR | 0.09 | 0.08 | 0.09 | 0.10 | 0.09 | 0.10 | SI | | 0.10 | 0.08 |
| Training on: PadChest | 0.05 | 0.05 | 0.02 | 0.02 | 0.05 | 0.06 | 0.17 | 0.19 | SI | |
| *Average* | *0.10* | *0.08* | *0.06* | *0.06* | *0.07* | *0.08* | *0.18* | *0.19* | *0.07* | *0.07* |
| P-value | 0.46 | 0.22 | 0.45 | 0.37 | 0.39 | 0.29 | 0.40 | 0.33 | 0.43 | 0.22 |

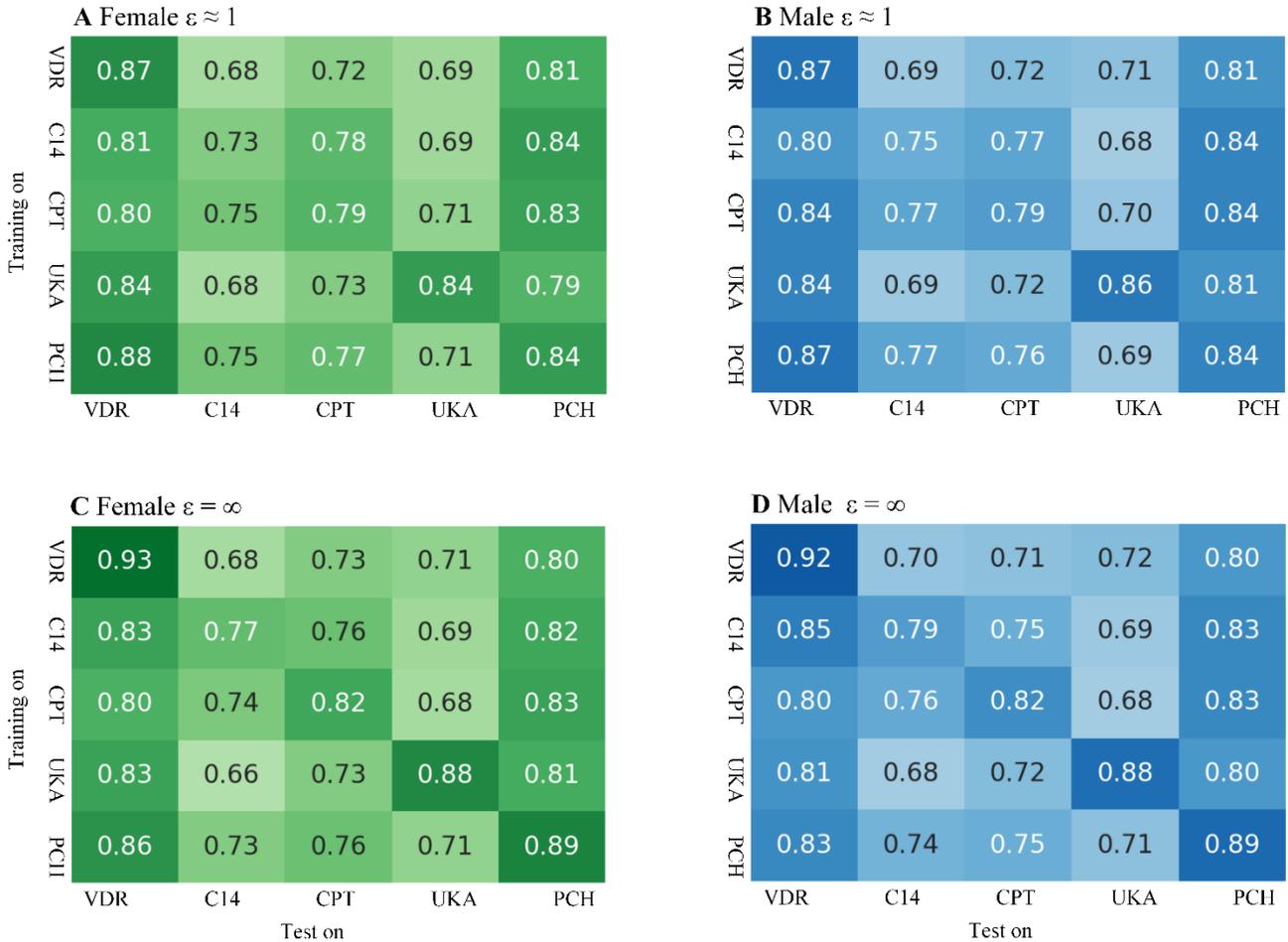

**Figure 4**: **Comparison between non-DP-DT (ε=∞) and DP-DT (ε≈1) in terms of average AUC for each sex subgroup.** Each row corresponds to a training domain, and each column corresponds to a test domain. The AUC values correspond to **(A)** female subgroups according to DP-DT with ε≈1, **(B)** male subgroups according to DP-DT with ε≈1, **(C)** female subgroups according to non-DP-DT (ε=∞), and **(D)** male subgroups according to non-DP-DT (ε=∞). AUC = area under the receiver operating characteristic curve, DP = differential privacy, DT = domain transfer, VDR=VinDr-CXR, C14=ChestX-ray14, CPT=CheXpert, UKA=UKA-CXR, PCH=PadChest.

## DP-DT Has No Effect on Age-Based Fairness

We repeated our experiments for different age-subgroups to similarly test if age-specific bias might be introduced. In **Figure 5**, we show the average AUC values individually for different age subgroups for every dataset evaluated on external domains both for DP-DT and non-DP-DT. The differences between these two methods are reported in **Table 4**. Except for the VinDr-CXR dataset with a maximum AUC difference of 0.04 (P=0.03 for individuals aged 70 to 100 years), which includes a small test sample size (n=149), we again observed on average, no evidence of differences, i.e., a maximum AUC difference of only 1% when comparing DP-DPT with non-DP-DT for all three age subgroups, including younger individuals (0-40 years), middle-aged individuals (40-70 years), and older individuals (70-100 years), in all datasets (P>0.25 for all cases). In a manner akin to the sex subgroups, more detailed age subgroup analyses are provided in **Tables S6–S9**.

**Table 4: Comparison of DP-Domain Transfer and non-DP-Domain Transfer Cross-Institutional Performance by Age Subgroup.** The differential values in each column correspond to AUC differences from the non-DP single-institutional case of that test dataset (baseline). The AUC results represent the average of all labels including cardiomegaly, pleural effusion, pneumonia, atelectasis, and healthy. The privacy budgets of the DP networks corresponding to each dataset are as follows: VinDr-CXR: $\varepsilon=1.17$, ChestX-ray14: $\varepsilon=1.01$, CheXpert: $\varepsilon=0.98$, UKA-CXR: $\varepsilon=0.98$, and PadChest: $\varepsilon=0.72$, with $\delta=0.000006$ for all datasets. P-values are calculated between DP and non-DP methods. Age subgroups are given in years.

| | Test on: VinDr-CXR | | | Test on: ChestX-ray14 | | | Test on: CheXpert | | | Test on: UKA-CXR | | | Test on: PadChest | | |
|---|---|---|---|---|---|---|---|---|---|---|---|---|---|---|---|
| | [0, 40) | [40, 70) | [70, 100) | [0, 40) | [40, 70) | [70, 100) | [0, 40) | [40, 70) | [70, 100) | [0, 40) | [40, 70) | [70, 100) | [0, 40) | [40, 70) | [70, 100) |
| $\varepsilon\approx1$ (DP) | | | | | | | | | | | | | | | |
| Training on: VinDr-CXR | SI | | | 0.08 | 0.08 | 0.08 | 0.09 | 0.09 | 0.10 | 0.17 | 0.16 | 0.17 | 0.06 | 0.09 | 0.09 |
| Training on: ChestX-ray14 | 0.05 | 0.07 | 0.08 | SI | | | 0.06 | 0.06 | 0.07 | 0.18 | 0.19 | 0.19 | 0.06 | 0.07 | 0.05 |
| Training on: CheXpert | 0.09 | 0.11 | 0.07 | 0.03 | 0.02 | 0.01 | SI | | | 0.19 | 0.20 | 0.20 | 0.04 | 0.06 | 0.06 |
| Training on: UKA-CXR | 0.07 | 0.07 | 0.09 | 0.11 | 0.10 | 0.11 | 0.09 | 0.09 | 0.10 | SI | | | 0.06 | 0.09 | 0.09 |
| Training on: PadChest | 0.05 | 0.08 | 0.08 | 0.04 | 0.03 | 0.03 | 0.07 | 0.06 | 0.07 | 0.15 | 0.17 | 0.18 | SI | | |
| *Average* | *0.07* | *0.08* | *0.08* | *0.07* | *0.06* | *0.06* | *0.08* | *0.08* | *0.09* | *0.17* | *0.18* | *0.19* | *0.06* | *0.08* | *0.07* |
| $\varepsilon=\infty$ (Non-DP) | | | | | | | | | | | | | | | |
| Training on: VinDr-CXR | SI | | | 0.09 | 0.08 | 0.10 | 0.10 | 0.10 | 0.11 | 0.17 | 0.17 | 0.19 | 0.07 | 0.08 | 0.08 |
| Training on: ChestX-ray14 | 0.08 | 0.08 | 0.01 | SI | | | 0.04 | 0.04 | 0.04 | 0.18 | 0.20 | 0.20 | 0.06 | 0.05 | 0.05 |
| Training on: CheXpert | 0.07 | 0.07 | 0.02 | 0.02 | 0.01 | 0.00 | SI | | | 0.15 | 0.18 | 0.18 | 0.05 | 0.06 | 0.05 |
| Training on: UKA-CXR | 0.08 | 0.11 | 0.09 | 0.09 | 0.09 | 0.09 | 0.10 | 0.09 | 0.09 | SI | | | 0.07 | 0.10 | 0.08 |
| Training on: PadChest | 0.03 | 0.05 | 0.03 | 0.02 | 0.01 | 0.01 | 0.04 | 0.04 | 0.06 | 0.16 | 0.19 | 0.19 | SI | | |
| *Average* | *0.07* | *0.08* | *0.04* | *0.06* | *0.05* | *0.05* | *0.07* | *0.07* | *0.08* | *0.17* | *0.19* | *0.19* | *0.06* | *0.07* | *0.07* |
| P-value | 0.50 | 0.38 | 0.03 | 0.36 | 0.37 | 0.42 | 0.35 | 0.35 | 0.30 | 0.25 | 0.33 | 0.27 | 0.16 | 0.36 | 0.30 |

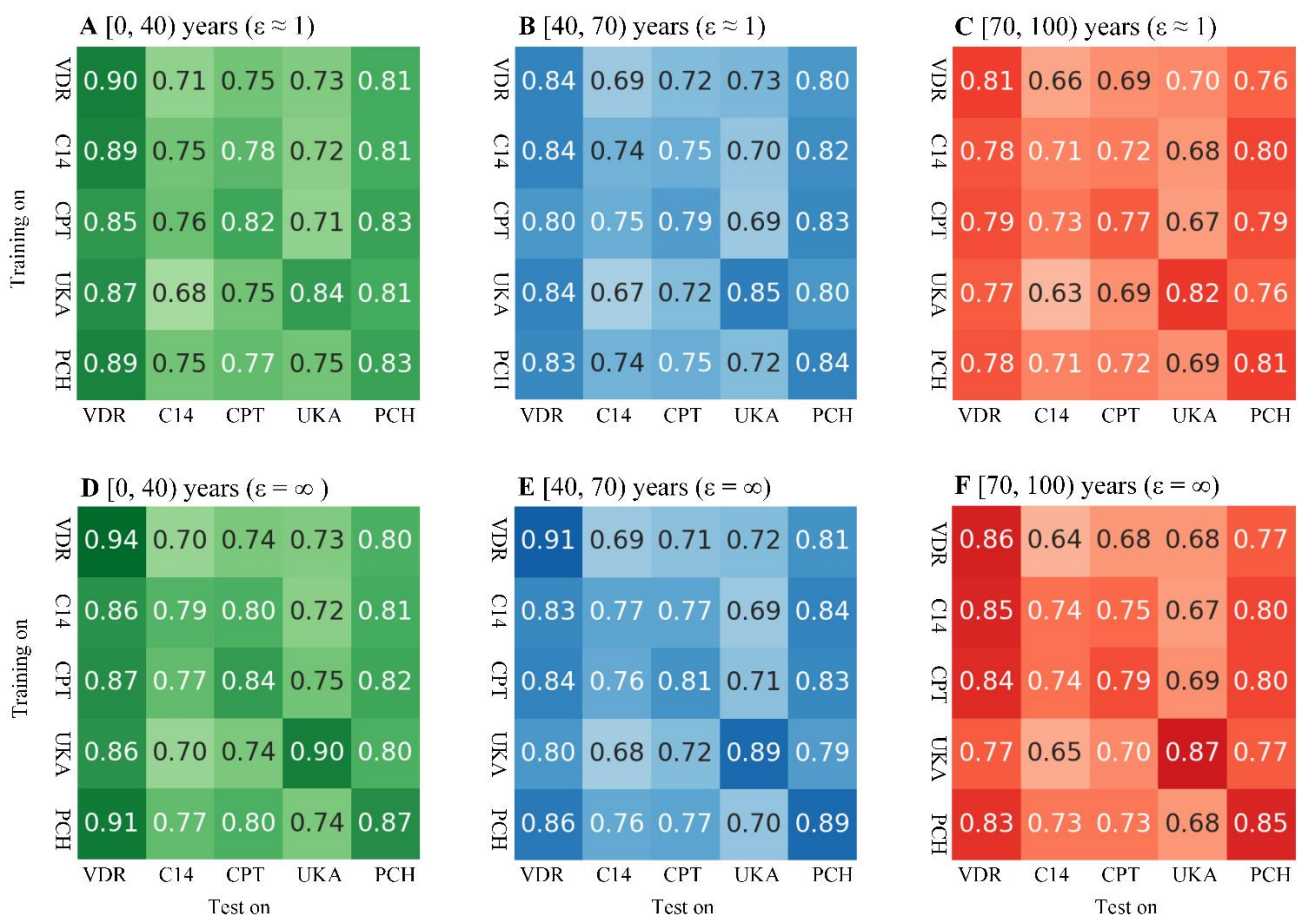

**Figure 5: Comparison between non-DP-DT (ε=∞) and DP-DT (ε≈1) in terms of average AUC for each age subgroup.** Each row corresponds to a training domain, and each column corresponds to a test domain. The AUC values correspond to **(A)** [0, 40) years subgroups according to DP-DT with ε≈1, **(B)** [40, 70) years subgroups according to DP-DT with ε≈1, **(C)** [70, 100) years subgroups according to DP-DT with ε≈1, **(D)** [0, 40) years subgroups according to non-DP-DT (ε=∞), **(E)** [40, 70) years subgroups according to non-DP-DT (ε=∞), and **(F)** [70, 100) years subgroups according to non-DP-DT (ε=∞). AUC = area under the receiver operating characteristic curve, DP = differential privacy, DT = domain transfer, VDR=VinDr-CXR, C14=ChestX-ray14, CPT=CheXpert, UKA=UKA-CXR, PCH=PadChest.

# Discussion

In this study, we investigated the domain transferability of highly privacy-preserving AI models in radiology for healthy patients and for patients diagnosed with cardiomegaly, pleural effusion, pneumonia, and atelectasis. Our analysis included a total of n=591,682 frontal chest radiographs from five different datasets from Vietnam, USA, Germany, and Spain, encompassing various imaging and labeling domains, such as standard upright imaging and intensive care imaging. As a baseline, we compared the performance of DP-trained networks in external domains with that of non-DP-trained networks. We used DP as the privacy-preserving technology to protect the private networks. Along with the comparison of network performance against the baseline, we conducted an analysis of the effects of DP on the fairness of the AI models when applied to different demographic sub-groups. This is a known issue when employing DP in deep learning models[11,16,25]. Our analysis aimed to provide insight into the potential trade-offs between privacy preservation and fairness as well as accuracy in AI models for medical diagnosis[11,17]. Our results indicate that at all privacy budgets - even with comparatively strict ε values around one - all DP models trained on any of the included datasets performed similarly to their non-DP counterparts and no evidence of differences were found in terms of average AUC when evaluated on external domains ($P>0.12$ for all cases). However, discrepancies between DP-trained and non-DP trained models were sometimes observed when testing for individual diseases; however, no consistent trend in favor of either training paradigm was observed.

Previous work[11] demonstrated that increasing the $\varepsilon$ value improves the diagnostic performance of a DP-trained AI model when tested on data from its own domain. Conversely, we show here that the cross-institutional performance of DP-trained models remains unaffected by increasing the $\varepsilon$ value. This is important, since any AI model in clinical practice will basically be acting as a cross-institutional model. Thus, our setup is more reflective of the clinical situation. Previous research[11] has emphasized the importance of large, curated training datasets for the successful generalization of DP-trained AI models. Interestingly, our findings suggest that this might not always be the paramount factor for cross-institutional applications. For instance, the VinDr-CXR dataset, comprising only n=15,000 training images, performed comparably to other datasets of varying sizes (ChestX-ray14: n=86,524, CheXpert: n=128,356, UKA-CXR: n=153,537, and PadChest: n=88,480) when juxtaposed with their non-DP counterpart for cross-institutional applications. We ascribe this observation to the inherent propensity of training with DP that mitigates overfitting during the training process.

We further investigated this finding for individual diseases, discovering that, except for cardiomegaly, the same pattern held for all individual diseases as well as healthy patients ($P>0.51$ for all cases). Even for cardiomegaly, with $P=0.01$, the AUC decrease was a mere 2.41%. Additionally, this finding was consistent across individual sex groups. We found that employing DP did not introduce any change in demographic parity in cross-institutional performance of the networks for both female and male subsets compared with non-DP-DT in terms of average AUC across all labels ($P>0.22$ for all cases).

Lastly, except for only one age group subset of the VinDr-CXR dataset (i.e., [70, 100) years), all age groups from all datasets followed the same trend, where no evidence of differences were found ($P>0.16$ for all cases). A closer examination of the age group, [70, 100), in the VinDr-CXR dataset revealed that this test subset was a small and underrepresented group with only 32 cardiomegaly, 4 pleural effusion, 9 pneumonia, 3 atelectasis, and 4 healthy samples. Thus, statistical fluctuations as the reason for this outlier are likely.

We recognize that attaining training convergence in DP AI models presents a more challenging and computationally intensive endeavor[11,14,15]. Nevertheless, by providing access to our comprehensive framework and recommended configurations, we aspire to expedite progress in this research area. In our experiments, with consistent computational resources, the DP training took on average 10 times longer to converge, in terms of total training time, than non-DP training, depending on the network architecture and dataset. A detailed computational efficiency analysis is provided in the supplementary materials. To the best of our knowledge, our study is the first to demonstrate this practical fact using various small and large real-world datasets from different domains.

While our study covers a wide range of real-world datasets, the main findings focus on the interpretation of chest radiographs using ResNet9. To broaden our results, supplemental materials present an ablation study featuring two additional network architectures[26,27] and more imaging findings. Here, a trend similar to our main results was evident. In future work, we intend to apply DP-DT across varied domains like gigapixel imaging in pathology, 3-dimensional volumetric medical imaging, and also more complex tasks, such as segmentation.

In summary, we conducted a comprehensive analysis of the application of DP in domain transferability of AI models based on chest radiographs, with our results demonstrating that employing DP in the training of diagnostic medical AI models does not impact the model's diagnostic performance and demographic fairness in external domains. Accordingly, we advocate for researchers and practitioners to place heightened priority on the integration of DP when training diagnostic medical AI models intended for collaborative applications. The present study affirms that even with extremely high privacy levels ($\varepsilon \approx 1$), DP does not cause minor trade-offs but exhibits nearly no impact on the network performance in external domains. We envisage that our observations will streamline collaborations among institutions, foster the development of more precise diagnostic AI models, and ultimately enhance patient outcomes.

# Materials and Methods

## Ethics Statement

This retrospective study was performed in accordance with relevant local and national guidelines and regulations and approved by the Ethical Committee of the Medical Faculty of RWTH Aachen University (Reference No. EK 028/19). The requirement to obtain individual informed consent was waived.

## Patient Cohorts

A total of 591,682 frontal chest radiographs from multiple institutions were included, comprising the VinDr-CXR[20] dataset with n=18,000, ChestX-ray14[21] dataset with n=112,120, CheXpert[22] dataset with n=157,676, UKA-CXR[9,11,23] dataset with n=193,361, and PadChest[24] dataset with n=110,525 radiographs. Since one patient might have multiple radiographs, we calculated privacy values per image.

Median and mean age over all patients were 61 and 59 ± [SD] 18 years, respectively, with a range from 1 to 111 years. **Table 1** reports the statistics of each dataset including labeling systems, age and sex distributions, and label distributions. Additionally, **Figures S1–S6** provide further information on the distribution of sample sizes per label and demographic subgroup for each of the datasets.

## Experimental Design

Two distinct networks, specifically, models trained employing DP- or non-DP training, were trained on a single dataset. Subsequently, testing was performed on a separate held-out test set of the same dataset (single-institutional) and corresponding test sets from the remaining datasets (cross-institutional), separately for both networks, resulting in single-institutional and cross-institutional performances for both DP-DT and non-DP-DT scenarios. Consequently, the held-out test sets were used for single-institutional and cross-institutional testing and comprised n=3,000 (VinDr-CXR), n=25,596 (ChestXray14), n=29,320 (CheXpert), n=39,824 (UKA-CXR), and n=22,045 (PadChest) chest radiographs (**Table 1**). Official test sets were employed for the VinDr-CXR and ChestX-ray14 datasets. Since no official test sets were available for the CheXpert, UKA-CXR, and PadChest datasets, images were randomly divided into 80% training and 20% test sets using a normal distribution. This division was patient-centric, ensuring all radiographs from one patient were grouped together, safeguarding patient-specific integrity and reducing potential underestimation of variance. The same training and test sets were used for both DP and Non-DP scenarios, making our comparisons intrinsically paired. It should be noted that we used a multilabel classification approach, optimizing for average performance across all labels and did not perform a detailed comparison for individual diseases.

**Harmonization of Labeling Systems**

In this study, the target labels for diagnosis were cardiomegaly, pleural effusion, pneumonia, and atelectasis. Additionally, we introduced a 'healthy' label for individuals who were not diagnosed with any pathology as recognized by the original datasets. A binary multilabel classification system was employed, meaning that each image could be diagnosed as either positive or negative for every disease. As a result, labels in datasets with non-binary labeling systems were converted to binary ones. Specifically, for datasets with certainty levels in labels (CheXpert), the "certain negative" and "uncertain" classes were considered negative labels, while only the "certain positive" class was counted as a positive label. For datasets with severity levels in labels (UKA-CXR), the threshold for differentiating between negative and positive labels was chosen as the middle of the severity levels. Lastly, in datasets with individual labels for each side of the body (UKA-CXR), both right and left labels for each disease were merged to form a single label per disease, meaning that the presence of a disease in at least one side was counted as positive.

**Privacy-Performance and Privacy-Fairness Trade-offs**

The privacy-performance trade-off was measured by analyzing each model's diagnostic performance using the area under the receiver operating characteristic curve (AUC) as the primary evaluation metric, with accuracy, sensitivity, and specificity as supporting evaluation metrics. The privacy-fairness trade-off was assessed by considering different demographics and subgroups within each dataset. We assumed that the network would be fair if it did not discriminate against any patient subgroups when introducing DP, meaning it should have the same performance in diagnosing any patient subgroups with and without DP. In addition to comparing the diagnostic performance in terms of AUC among different subgroups, the statistical parity difference[28] was further used for demographic fairness analysis. Demographic subgroups considered in this study included females, males, and patients within the age ranges of [0, 40), [40, 70), and [70, 100) years.

# Image Pre-Processing

As described in previous studies[9,11], a unified image pre-processing strategy was applied to all datasets, which included: i) resizing all images to a size of $512 \times 512$, ii) performing min-max normalization (feature scaling) as proposed by et al.[29], and iv) performing histogram equalization.

# DL Network Architecture and Training

To ensure compatibility with DP training, a modified ResNet9 architecture was used, incorporating modifications proposed by Klause et al.[30] and by He et al.[26]. Group normalization[31] with groups of 32 was used instead of batch normalization[32]. Mish[33] was selected as the activation function. The inputs to the network were 3-channel images, with the output of the first layer having 64 channels. Finally, a fully connected layer reduced the 512 features to 5. The logistic sigmoid function was employed for converting output predictions to class probabilities.

    Previous work[11] demonstrated that appropriate pre-training is essential for the convergence of medical DP models. Consequently, our network was pre-trained on the publicly available MIMIC-CXR[29] dataset, consisting of n=210,652 frontal chest radiographs. All models were optimized using the NAdam optimizer, with learning rates ranging from $1 \times 10^{-4}$ to $5 \times 10^{-4}$ for optimal convergence, and no weight decay applied. Binary cross-entropy was chosen as the loss function. Data augmentation during non-DP training was implemented by applying random rotation within the range of [-8, 8] degrees and flipping[9]. In contrast, no data augmentation was performed during DP training due to its reported negative effect[11]. The maximum allowed gradient norm was found to be an influential factor in DP network convergence, with an optimal value of 1.5 observed for all trainings. Each point in the DP training batches was sampled with a probability of 128 divided by the sample size for each dataset, while a batch size of 128 was used in non-DP training. A DP accountant employing Rényi differential privacy[34] was chosen. This mechanism oversees the 'privacy budget' (represented by ε and δ) and ensures its adherence to predetermined bounds. $\delta = 6 \times 10^{-6}$ was

selected for all datasets[11]. ε depends on the introduced noise, the set δ, and factors like training steps and batch size. Given the dataset diversity, each neural network's convergence step determined the reported ε.

## Evaluation Metrics and Statistical Analysis

Statistical analysis was carried out using Python (v3) and the associated packages SciPy, and NumPy. We employed the AUC as our primary evaluation metric. Individual label results were averaged without weighting. Accuracy, sensitivity, and specificity across varied demographics were calculated as secondary evaluation metrics. Bootstrapping[35] was applied to each test set to assess the statistical spread, with 1,000 redraws. In each redraw, radiographs were randomly picked, matching the set size, and this selection allowed repetitions. To assess statistical significance between evaluation metrics obtained with DP compared with those obtained with non-DP, a two-tailed Student's t-test was used. Multiplicity-adjusted p-values were determined based on the false discovery rate to account for multiple comparisons, and the family-wise alpha threshold was set at 0.05.

## Code Availability

All source codes for training and evaluation of the deep neural networks, data augmentation, image analysis, and preprocessing are publicly available at https://github.com/tayebiarasteh/privacydomain. All code for the experiments was developed in Python v3.10 using the PyTorch v1.13 framework. The differential privacy was developed using Opacus[36] v1.3.0.

## Data Availability

The accessibility of the utilized data in this study is as follows: ChestX-ray14 and PadChest datasets are publicly available via https://www.v7labs.com/open-datasets/chestx-ray14 and https://bimcv.cipf.es/bimcv-projects/padchest/, respectively. VinDr-CXR and MIMIC-CXR datasets are restricted-access resources, which can be accessed from PhysioNet by agreeing to its data protection requirements under https://physionet.org/content/vindr-cxr/1.0.0/ and https://physionet.org/content/mimic-cxr-jpg/2.0.0/, respectively. CheXpert data could be requested from Stanford University at https://stanfordmlgroup.github.io/competitions/chexpert/. The UKA-CXR data is not publicly accessible as it is internal data of patients of University Hospital RWTH Aachen in Aachen, Germany. Data access can be granted upon reasonable request to the corresponding author.

## Hardware

The hardware used in our experiments were Intel CPUs with 18 cores and 32 GB RAM and Nvidia RTX 6000 GPUs with 24 GB memory.

# Additional information

### Funding Sources

This work was partially funded and supported by the Radiological Cooperative Network (RACOON) under BMBF grant number 01KX2021 and has been funded by the German Federal Ministry of Education and Research and the Bavarian State Ministry for Science and the Arts. The authors of this work take full responsibility for its content.

### Author Contributions

STA, GK, and DT designed the study. The manuscript was written by STA and reviewed and corrected by TN, GK, and DT. The experiments were performed by STA. The software was developed by STA. Illustrations were designed by STA, ML, and TN. The statistical analyses were performed by STA, ML, SN, GK, and DT. STA preprocessed the data. MS, PI, CK, SN, DT, and GK provided clinical expertise. All authors read the manuscript and agreed to the submission of this paper.

### Competing Interests

The authors do not have any competing interest to disclose.

### Correspondence:

Soroosh Tayebi Arasteh, MSc

Department of Diagnostic and Interventional Radiology, University Hospital RWTH Aachen, Pauwelsstr. 30, 52074 Aachen, Germany
Email: sarasteh@ukaachen.de

# Supplemental Material

# Appendix S1

### Further Statistics of Patient Cohorts

**Figures S1–S6** show total sample sizes per individual positive labels of the full dataset as well as for different demographic subsets of the respective test dataset. VDR=VinDr-CXR[20], C14=ChestX-ray14[21], CPT=CheXpert[22], UKA=UKA-CXR[9,11,23], PCH=PadChest[24], CM=cardiomegaly, PE=pleural effusion, PN=pneumonia, AT=atelectasis, HE=healthy, AI=artificial intelligence, AUC=area under the receiver operating characteristic curve, DP=differential privacy, DP-DT=differential privacy-enhanced domain transfer, DT=domain transfer.

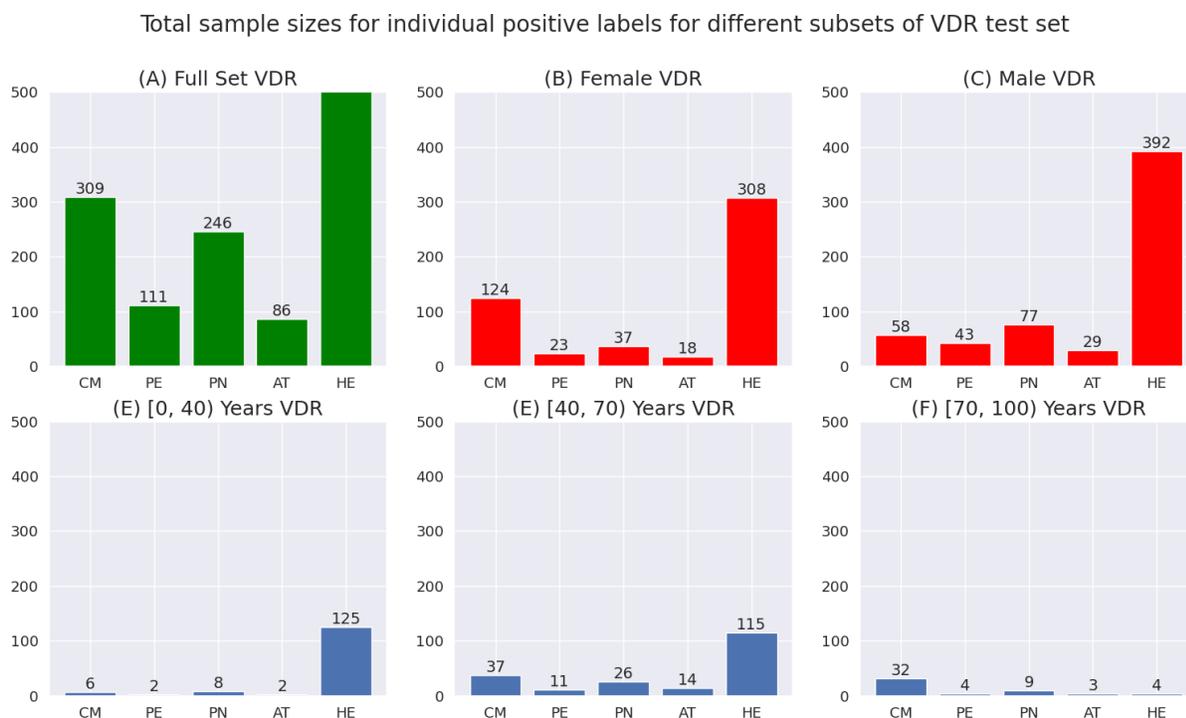

**Figure S1: Total sample sizes for individual positive labels for different subsets of the VinDr-CXR test set. (A)** full test set, **(B)** females, **(C)** males, **(D)** age group between [0, 40) years, **(E)** age group between [40, 70) years, and **(F)** age group between [70, 100) years. Note that not all of the subject ages were available for this dataset. VDR=VinDr-CXR, C14=ChestX-ray14, CPT=CheXpert, UKA=UKA-CXR, PCH=PadChest, CM=cardiomegaly, PE=pleural effusion, PN=pneumonia, AT=atelectasis, HE=healthy.

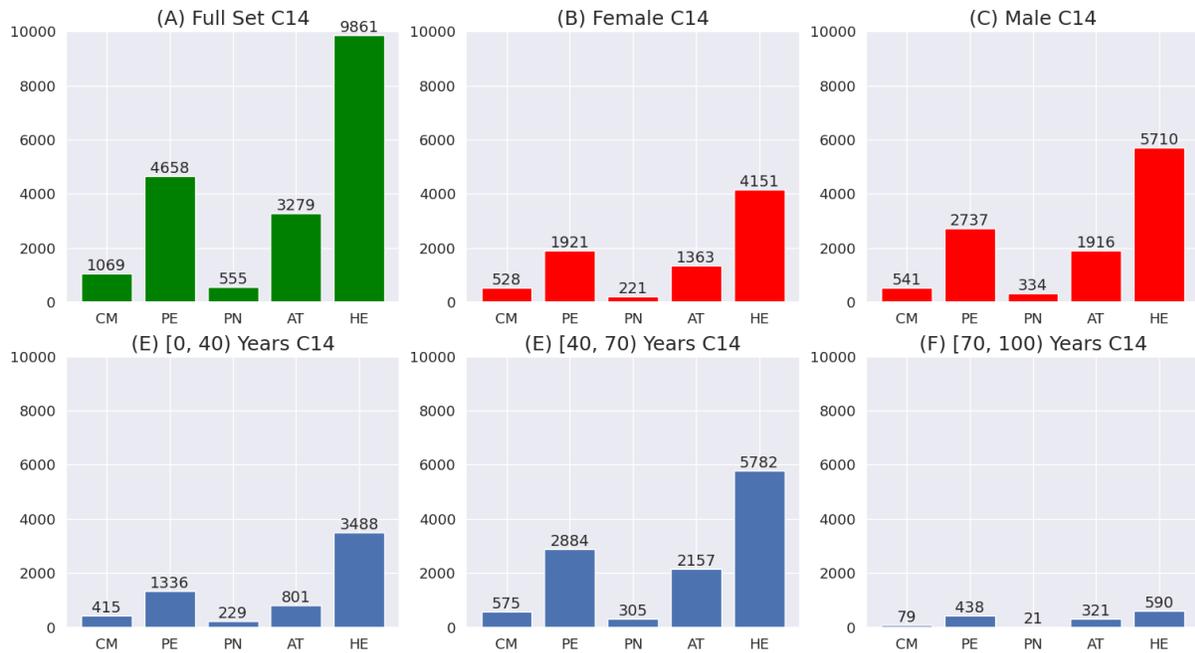

**Figure S2: Total sample sizes for individual positive labels for different subsets of the ChestX-ray14 test set.** **(A)** full test set, **(B)** females, **(C)** males, **(D)** age group between [0, 40) years, **(E)** age group between [40, 70) years, and **(F)** age group between [70, 100) years. VDR=VinDr-CXR, C14=ChestX-ray14, CPT=CheXpert, UKA=UKA-CXR, PCH=PadChest, CM=cardiomegaly, PE=pleural effusion, PN=pneumonia, AT=atelectasis, HE=healthy.

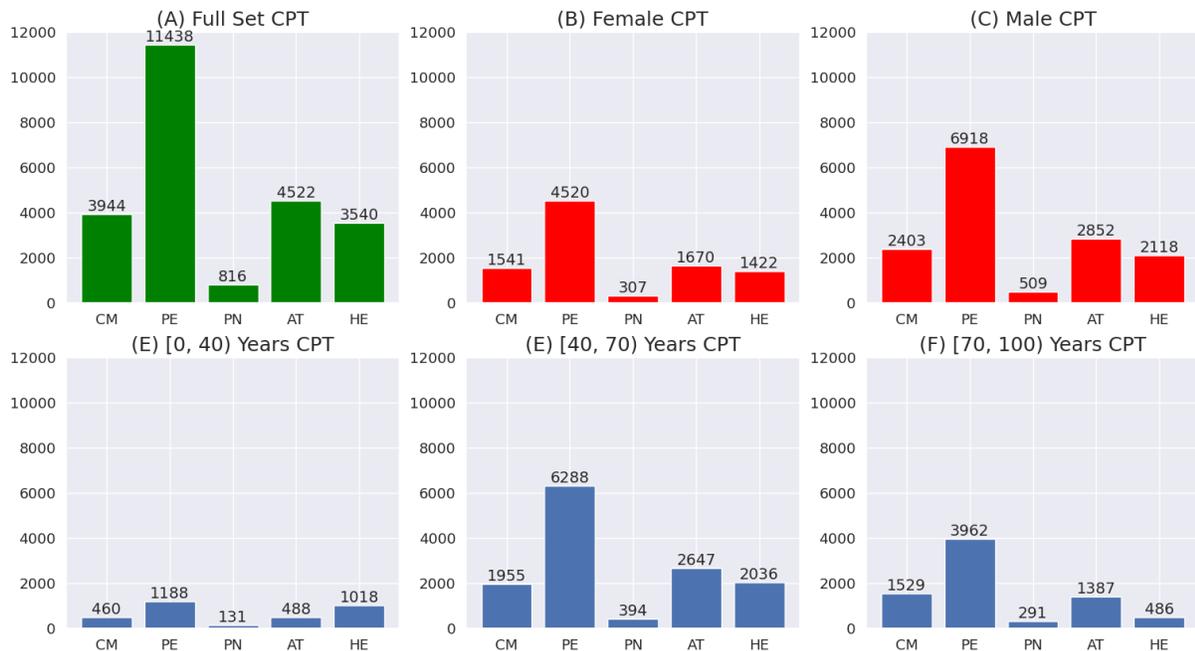

**Figure S3: Total sample sizes for individual positive labels for different subsets of the CheXpert test set. (A)** full test set, **(B)** females, **(C)** males, **(D)** age group between [0, 40) years, **(E)** age group between [40, 70) years, and **(F)** age group between [70, 100) years. VDR=VinDr-CXR, C14=ChestX-ray14, CPT=CheXpert, UKA=UKA-CXR, PCH=PadChest, CM=cardiomegaly, PE=pleural effusion, PN=pneumonia, AT=atelectasis, HE=healthy.

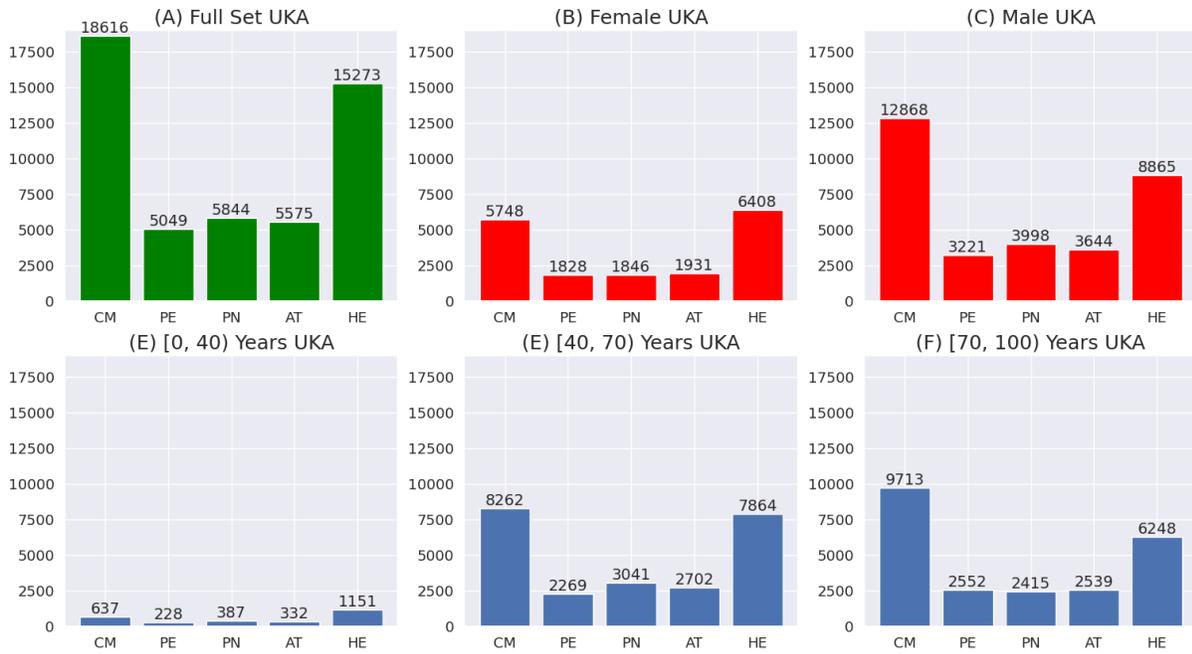

**Figure S4: Total sample sizes for individual positive labels for different subsets of the UKA-CXR test set. (A)** full test set, **(B)** females, **(C)** males, **(D)** age group between [0, 40) years, **(E)** age group between [40, 70) years, and **(F)** age group between [70, 100) years. VDR=VinDr-CXR, C14=ChestX-ray14, CPT=CheXpert, UKA=UKA-CXR, PCH=PadChest, CM=cardiomegaly, PE=pleural effusion, PN=pneumonia, AT=atelectasis, HE=healthy.

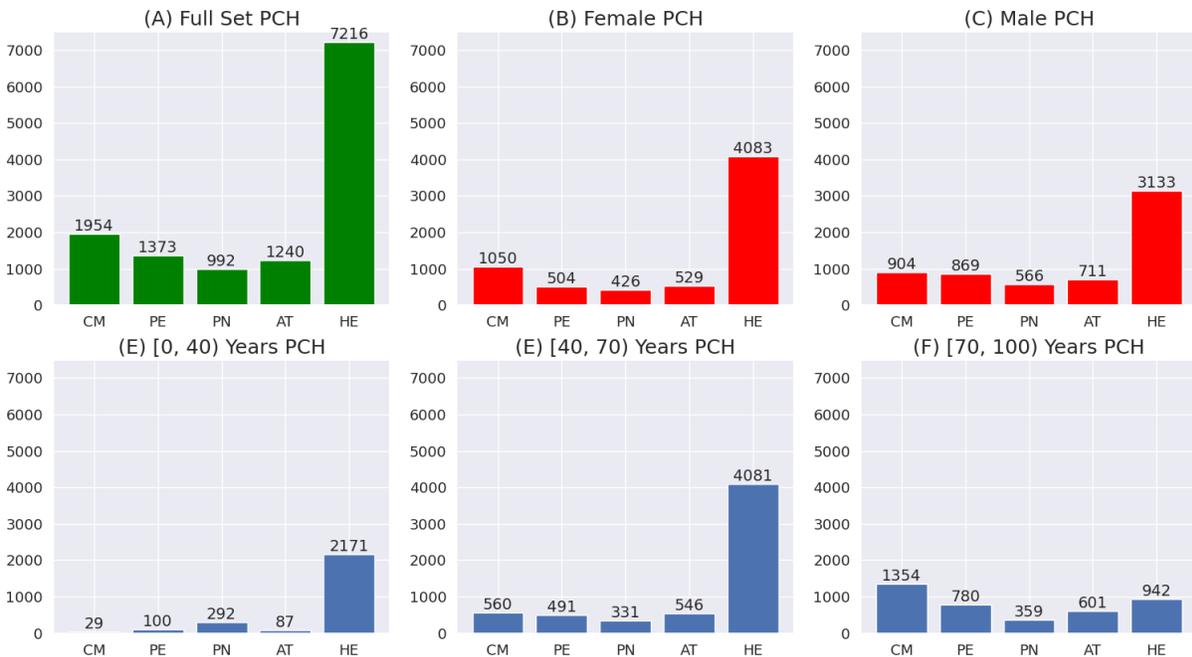

**Figure S5: Total sample sizes for individual positive labels for different subsets of the PadChest test set. (A)** full test set, **(B)** females, **(C)** males, **(D)** age group between [0, 40) years, **(E)** age group between [40, 70) years, and **(F)** age group between [70, 100) years. VDR=VinDr-CXR, C14=ChestX-ray14, CPT=CheXpert, UKA=UKA-CXR, PCH=PadChest, CM=cardiomegaly, PE=pleural effusion, PN=pneumonia, AT=atelectasis, HE=healthy.

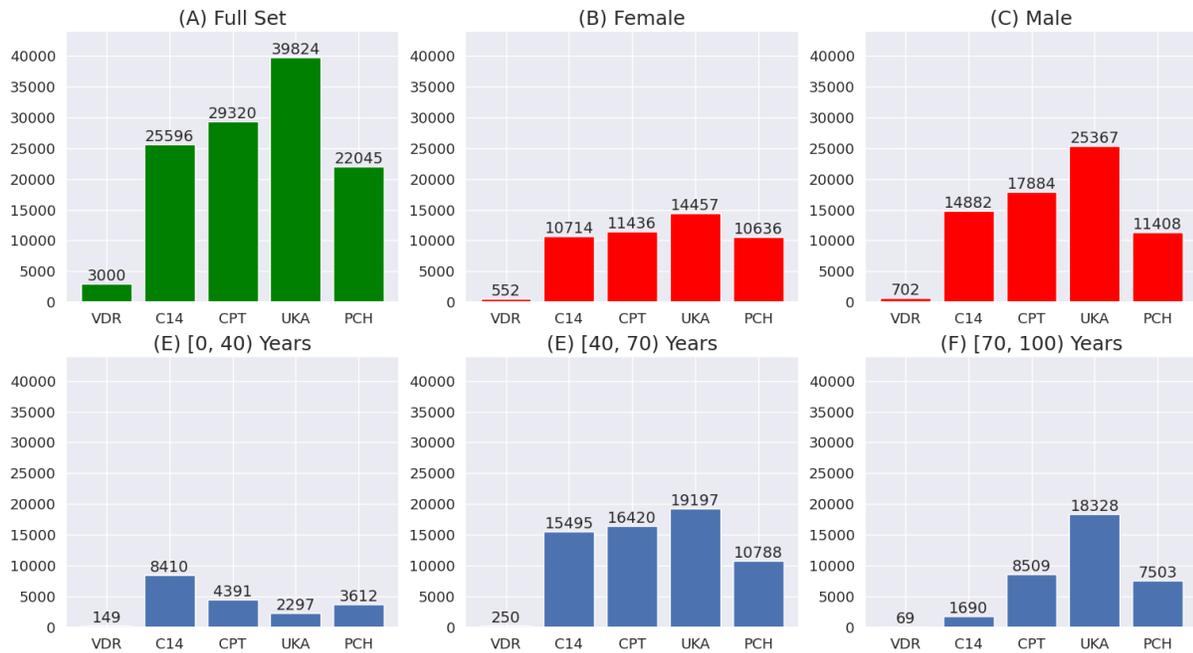

**Figure S6: Overall sample sizes for different subsets of each test benchmark. (A)** full test set, **(B)** females, **(C)** males, **(D)** age group between [0, 40) years, **(E)** age group between [40, 70) years, and **(F)** age group between [70, 100) years. VDR=VinDr-CXR, C14=ChestX-ray14, CPT=CheXpert, UKA=UKA-CXR, PCH=PadChest.

## Area Under the Receiver Operating Characteristic Curve Figures For Other ε Values

**Figures S7–S9** show the results of transferring non-DP models to different domains including VinDr-CXR[20] (VDR): n=15,000, ChestX-ray14[21] (C14): n=86,524, CheXpert[22] (CPT): n=128,356, UKA-CXR[9,11,23] (UKA): n=153,537, and PadChest[24] (PCH): n=88,480, for individual labels. δ=0.000006 was chosen for all datasets.

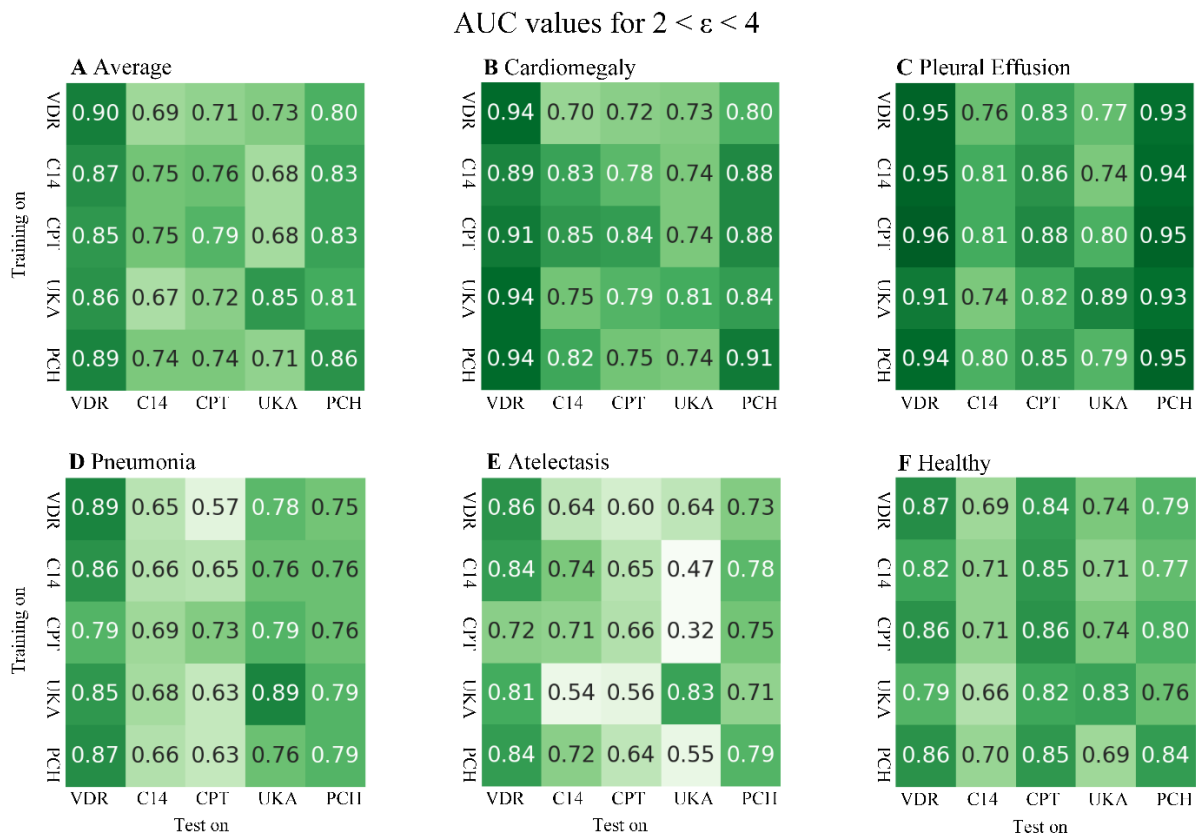

**Figure S7: Results of transferring differential privacy (DP) models with 2 < ε < 4 to different domains for individual labels.** The area under the receiver operating characteristic curve (AUC) values correspond to **(A)** average over all labels, **(B)** cardiomegaly, **(C)** pleural effusion, **(D)** pneumonia, **(E)** atelectasis, and **(F)** healthy. Each row corresponds to a training domain and each column corresponds to a test domain. The privacy budgets of the DP networks corresponding to each dataset are as follows: VinDr-CXR (VDR): $\varepsilon = 3.24$, ChestX-ray14 (C14): $\varepsilon = 3.37$, CheXpert (CPT): $\varepsilon = 3.30$, UKA-CXR (UKA): $\varepsilon = 3.45$, and PadChest (PCH): $\varepsilon = 3.58$.

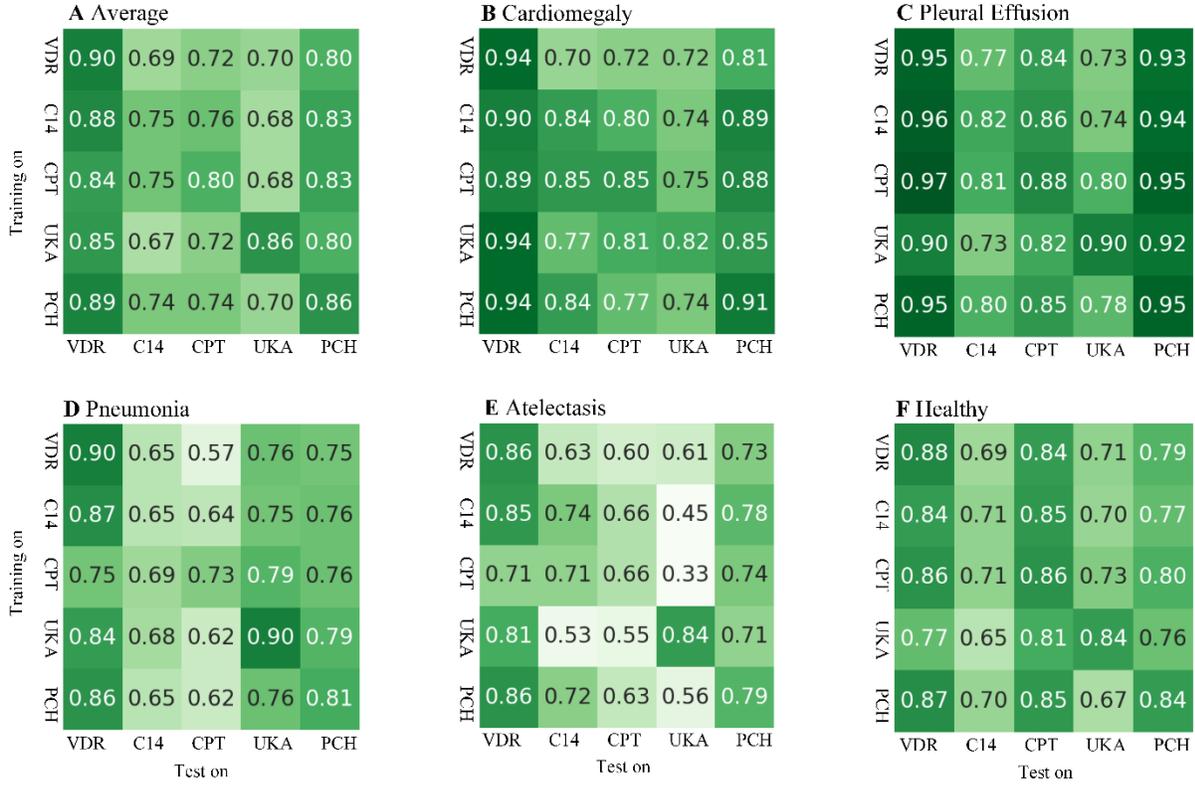

**Figure S8: Results of transferring differential privacy (DP) models with 4 < ε < 9 to different domains for individual labels.** The area under the receiver operating characteristic curve (AUC) values correspond to **(A)** average over all labels, **(B)** cardiomegaly, **(C)** pleural effusion, **(D)** pneumonia, **(E)** atelectasis, and **(F)** healthy. Each row corresponds to a training domain and each column corresponds to a test domain. The privacy budgets of the DP networks corresponding to each dataset are as follows VinDr-CXR (VDR): $\varepsilon = 4.29$, ChestX-ray14 (C14): $\varepsilon = 7.83$, CheXpert (CPT): $\varepsilon = 6.48$, UKA-CXR (UKA): $\varepsilon = 8.81$, and PadChest (PCH): $\varepsilon = 7.41$.

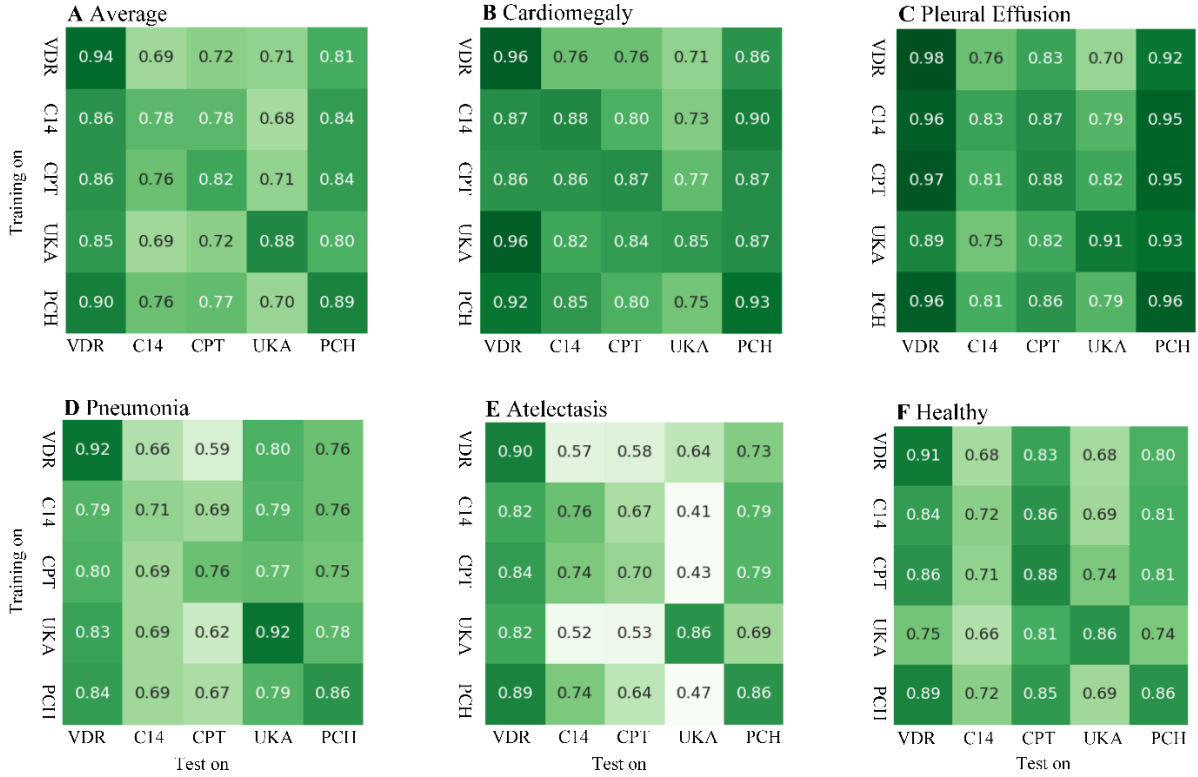

**Figure S9: Results of transferring non-differential privacy (DP) models (ε=∞) to different domains for individual labels.** The area under the receiver operating characteristic curve (AUC) values correspond to **(A)** average over all labels, **(B)** cardiomegaly, **(C)** pleural effusion, **(D)** pneumonia, **(E)** atelectasis, and **(F)** healthy. Each row corresponds to a training domain and each column corresponds to a test domain. VDR=VinDr-CXR, C14=ChestX-ray14, CPT=CheXpert, UKA=UKA-CXR, PCH=PadChest.

## Detailed Evaluation Results of Experiments

The area under the receiver operating characteristic curve (AUC) values for individual labels and further evaluation metrics including accuracy, sensitivity, specificity, and statistical parity difference are provided for all experiments averaged over all labels (cardiomegaly, pleural effusion, pneumonia, atelectasis, and healthy) in **Tables S1–S9**.
26

**Table S1: More detailed comparison of Differential Privacy-enhanced Domain Transfer (DP-DT) and non-DP-DT in terms of cross-institutional performance.** Here we present the accuracy (ACC), sensitivity (Sens), and specificity (Spec) values for each training-testing combination, averaged over all one-vs-all classifications (cardiomegaly, pleural effusion, pneumonia, atelectasis, and healthy). Note that there is no significant difference between the DP-trained and the non-DP-trained networks. The privacy budgets of the DP networks are as follows: VinDr-CXR: $\varepsilon=1.17$, ChestX-ray14: $\varepsilon=1.01$, CheXpert: $\varepsilon=0.98$, UKA-CXR: $\varepsilon=0.98$, and PadChest: $\varepsilon=0.72$, with $\delta = 0.000006$ for all datasets. P-values were calculated between DP and non-DP methods (for cross-institutional only).

| | Overall | | | Test on: VinDr-CXR | | | Test on: ChestX-ray14 | | | Test on: CheXpert | | | Test on: UKA-CXR | | | Test on: PadChest | | |
|---|---|---|---|---|---|---|---|---|---|---|---|---|---|---|---|---|---|---|
| | ACC | Sens | Spec | ACC | Sens | Spec | ACC | Sens | Spec | ACC | Sens | Spec | ACC | Sens | Spec | ACC | Sens | Spec |
| $\varepsilon \approx 1$ (DP) | | | | | | | | | | | | | | | | | | |
| Training on: VinDr-CXR | 0.82 | 0.84 | 0.81 | 0.57 | 0.72 | 0.58 | 0.66 | 0.71 | 0.63 | 0.64 | 0.70 | 0.63 | 0.73 | 0.78 | 0.72 | | | |
| Training on: ChestX-ray14 | 0.82 | 0.81 | 0.81 | 0.67 | 0.69 | 0.68 | 0.68 | 0.74 | 0.66 | 0.59 | 0.74 | 0.55 | 0.74 | 0.80 | 0.73 | | | |
| Training on: CheXpert | 0.78 | 0.78 | 0.79 | 0.68 | 0.70 | 0.69 | 0.73 | 0.75 | 0.72 | 0.59 | 0.76 | 0.56 | 0.74 | 0.81 | 0.73 | | | |
| Training on: UKA-CXR | 0.81 | 0.78 | 0.81 | 0.56 | 0.72 | 0.56 | 0.64 | 0.74 | 0.62 | 0.75 | 0.78 | 0.75 | 0.73 | 0.80 | 0.71 | | | |
| Training on: PadChest | 0.82 | 0.79 | 0.83 | 0.67 | 0.69 | 0.68 | 0.67 | 0.73 | 0.66 | 0.65 | 0.67 | 0.65 | 0.77 | 0.79 | 0.76 | | | |
| *Cross-institutional average* | *0.81* | *0.79* | *0.81* | *0.62* | *0.71* | *0.63* | *0.66* | *0.73* | *0.64* | *0.62* | *0.72* | *0.60* | *0.74* | *0.80* | *0.72* | | | |
| $\varepsilon = \infty$ (Non-DP) | | | | | | | | | | | | | | | | | | |
| Training on: VinDr-CXR | 0.88 | 0.88 | 0.87 | 0.59 | 0.71 | 0.59 | 0.66 | 0.70 | 0.64 | 0.63 | 0.69 | 0.62 | 0.72 | 0.80 | 0.71 | | | |
| Training on: ChestX-ray14 | 0.80 | 0.78 | 0.81 | 0.71 | 0.70 | 0.72 | 0.70 | 0.75 | 0.69 | 0.71 | 0.58 | 0.73 | 0.76 | 0.80 | 0.75 | | | |
| Training on: CheXpert | 0.81 | 0.79 | 0.80 | 0.68 | 0.71 | 0.69 | 0.73 | 0.78 | 0.72 | 0.60 | 0.76 | 0.58 | 0.76 | 0.79 | 0.76 | | | |
| Training on: UKA-CXR | 0.82 | 0.79 | 0.79 | 0.62 | 0.68 | 0.62 | 0.66 | 0.71 | 0.65 | 0.79 | 0.82 | 0.78 | 0.72 | 0.79 | 0.71 | | | |
| Training on: PadChest | 0.83 | 0.84 | 0.83 | 0.69 | 0.71 | 0.70 | 0.71 | 0.72 | 0.69 | 0.58 | 0.77 | 0.55 | 0.82 | 0.84 | 0.81 | | | |
| *Cross-institutional average* | *0.82* | *0.80* | *0.81* | *0.65* | *0.70* | *0.65* | *0.68* | *0.72* | *0.67* | *0.63* | *0.70* | *0.62* | *0.74* | *0.80* | *0.73* | | | |
| P-value | 0.52 | 0.58 | 0.72 | 0.14 | 0.73 | 0.19 | 0.09 | 0.31 | 0.01 | 0.77 | 0.77 | 0.73 | 0.60 | 0.79 | 0.35 | | | |

**Table S2: Average cross-institutional evaluation results for individual diseases.** The table presents results for two cases of DP with ε≈1 and non-DP networks. The values in each column show the average of all the cross-institutional AUC values when trained on that dataset. The privacy budgets of the DP networks corresponding to each dataset are as follows: VinDr-CXR: ε=1.17, ChestX-ray14: ε=1.01, CheXpert: ε=0.98, UKA-CXR: ε=0.98, and PadChest: ε=0.72, with $\delta = 0.000006$ for all datasets. P-values are calculated between DP and non-DP methods for every label.

|  | Cardiomegaly | Pleural Effusion | Pneumonia | Atelectasis | Healthy | *Average* |
|---|---|---|---|---|---|---|
| ε≈1 (DP) |  |  |  |  |  |  |
| Training on: VinDr-CXR | 0.74 | 0.83 | 0.70 | 0.65 | 0.76 | *0.74* |
| Training on: ChestX-ray14 | 0.82 | 0.87 | 0.78 | 0.69 | 0.80 | *0.79* |
| Training on: CheXpert | 0.84 | 0.88 | 0.76 | 0.64 | 0.77 | *0.76* |
| Training on: UKA-CXR | 0.83 | 0.85 | 0.74 | 0.66 | 0.76 | *0.75* |
| Training on: PadChest | 0.81 | 0.85 | 0.74 | 0.69 | 0.78 | *0.77* |
| *Average* | *0.81* | *0.85* | *0.74* | *0.67* | *0.77* | *0.76* |
| ε=∞ (Non-DP) |  |  |  |  |  |  |
| Training on: VinDr-CXR | 0.77 | 0.80 | 0.70 | 0.63 | 0.75 | *0.73* |
| Training on: ChestX-ray14 | 0.83 | 0.89 | 0.76 | 0.67 | 0.80 | *0.79* |
| Training on: CheXpert | 0.84 | 0.89 | 0.75 | 0.70 | 0.78 | *0.79* |
| Training on: UKA-CXR | 0.87 | 0.85 | 0.73 | 0.64 | 0.74 | *0.77* |
| Training on: PadChest | 0.83 | 0.86 | 0.75 | 0.69 | 0.79 | *0.78* |
| *Average* | *0.83* | *0.86* | *0.74* | *0.67* | *0.77* | *0.77* |
| P-value | 0.01 | 0.55 | 0.82 | 0.96 | 0.70 | 0.51 |

**Table S3: More detailed comparison of DP-DT and non-DP-DT in terms of cross-institutional performance for female subgroup.** Here we present the accuracy (ACC), sensitivity (Sens), and specificity (Spec) values for each training-testing combination, averaged over all one-vs-all classifications (cardiomegaly, pleural effusion, pneumonia, atelectasis, and healthy). Note that there is no significant difference between the DP-trained and the non-DP-trained networks. The privacy budgets of the DP networks are as follows: VinDr-CXR: ε=1.17, ChestX-ray14: ε=1.01, CheXpert: ε=0.98, UKA-CXR: ε=0.98, and PadChest: ε=0.72, with $\delta = 0.000006$ for all datasets. P-values were calculated between DP and non-DP methods (for cross-institutional only).

| Females | Test on: VinDr-CXR | | | Test on: ChestX-ray14 | | | Test on: CheXpert | | | Test on: UKA-CXR | | | Test on: PadChest | | |
|---|---|---|---|---|---|---|---|---|---|---|---|---|---|---|---|
| | ACC | Sens | Spec | ACC | Sens | Spec | ACC | Sens | Spec | ACC | Sens | Spec | ACC | Sens | Spec |
| ε≈1 (DP) | | | | | | | | | | | | | | | |
| Training on: VinDr-CXR | 0.82 | 0.76 | 0.83 | 0.57 | 0.70 | 0.58 | 0.66 | 0.73 | 0.63 | 0.64 | 0.69 | 0.63 | 0.76 | 0.75 | 0.75 |
| Training on: ChestX-ray14 | 0.79 | 0.74 | 0.80 | 0.66 | 0.69 | 0.67 | 0.67 | 0.75 | 0.66 | 0.61 | 0.70 | 0.59 | 0.77 | 0.78 | 0.76 |
| Training on: CheXpert | 0.76 | 0.70 | 0.78 | 0.68 | 0.68 | 0.69 | 0.73 | 0.74 | 0.72 | 0.63 | 0.69 | 0.62 | 0.76 | 0.79 | 0.75 |
| Training on: UKA-CXR | 0.78 | 0.77 | 0.79 | 0.56 | 0.71 | 0.56 | 0.65 | 0.74 | 0.62 | 0.75 | 0.79 | 0.73 | 0.75 | 0.78 | 0.74 |
| Training on: PadChest | 0.81 | 0.74 | 0.82 | 0.68 | 0.66 | 0.70 | 0.67 | 0.75 | 0.66 | 0.65 | 0.68 | 0.65 | 0.78 | 0.78 | 0.78 |
| *Cross-institutional average* | *0.79* | *0.74* | *0.80* | *0.62* | *0.69* | *0.63* | *0.66* | *0.74* | *0.64* | *0.63* | *0.69* | *0.62* | *0.76* | *0.78* | *0.75* |
| ε=∞ (Non-DP) | | | | | | | | | | | | | | | |
| Training on: VinDr-CXR | 0.90 | 0.84 | 0.90 | 0.60 | 0.68 | 0.61 | 0.65 | 0.72 | 0.63 | 0.64 | 0.66 | 0.63 | 0.75 | 0.77 | 0.74 |
| Training on: ChestX-ray14 | 0.75 | 0.74 | 0.77 | 0.71 | 0.69 | 0.72 | 0.70 | 0.76 | 0.68 | 0.71 | 0.57 | 0.73 | 0.77 | 0.79 | 0.76 |
| Training on: CheXpert | 0.75 | 0.72 | 0.78 | 0.67 | 0.71 | 0.68 | 0.73 | 0.79 | 0.72 | 0.61 | 0.74 | 0.60 | 0.77 | 0.79 | 0.76 |
| Training on: UKA-CXR | 0.82 | 0.75 | 0.83 | 0.62 | 0.66 | 0.63 | 0.64 | 0.75 | 0.61 | 0.79 | 0.82 | 0.78 | 0.73 | 0.77 | 0.71 |
| Training on: PadChest | 0.81 | 0.81 | 0.83 | 0.68 | 0.70 | 0.70 | 0.69 | 0.75 | 0.68 | 0.60 | 0.74 | 0.59 | 0.82 | 0.84 | 0.82 |
| *Cross-institutional average* | *0.78* | *0.76* | *0.80* | *0.64* | *0.69* | *0.66* | *0.67* | *0.75* | *0.65* | *0.64* | *0.68* | *0.64* | *0.76* | *0.78* | *0.74* |
| P-value | 0.89 | 0.43 | 0.75 | 0.29 | 0.99 | 0.30 | 0.52 | 0.64 | 0.39 | 0.83 | 0.79 | 0.75 | 0.49 | 0.49 | 0.44 |

**Table S4: More detailed comparison of DP-DT and non-DP-DT in terms of cross-institutional performance for male subgroup.** Here we present the accuracy (ACC), sensitivity (Sens), and specificity (Spec) values for each training-testing combination, averaged over all one-vs-all classifications (cardiomegaly, pleural effusion, pneumonia, atelectasis, and healthy). Note that there is no significant difference between the DP-trained and the non-DP-trained networks. The privacy budgets of the DP networks are as follows: VinDr-CXR: $\varepsilon=1.17$, ChestX-ray14: $\varepsilon=1.01$, CheXpert: $\varepsilon=0.98$, UKA-CXR: $\varepsilon=0.98$, and PadChest: $\varepsilon=0.72$, with $\delta = 0.000006$ for all datasets. P-values were calculated between DP and non-DP methods (for cross-institutional only).

| Males | Test on: VinDr-CXR | | | Test on: ChestX-ray14 | | | Test on: CheXpert | | | Test on: UKA-CXR | | | Test on: PadChest | | |
|---|---|---|---|---|---|---|---|---|---|---|---|---|---|---|---|
| | ACC | Sens | Spec | ACC | Sens | Spec | ACC | Sens | Spec | ACC | Sens | Spec | ACC | Sens | Spec |
| $\varepsilon\approx 1$ (DP) | | | | | | | | | | | | | | | |
| Training on: VinDr-CXR | 0.81 | 0.81 | 0.80 | 0.59 | 0.72 | 0.59 | 0.65 | 0.71 | 0.63 | 0.64 | 0.71 | 0.63 | 0.71 | 0.80 | 0.69 |
| Training on: ChestX-ray14 | 0.78 | 0.78 | 0.78 | 0.68 | 0.69 | 0.69 | 0.68 | 0.73 | 0.66 | 0.57 | 0.76 | 0.54 | 0.73 | 0.82 | 0.71 |
| Training on: CheXpert | 0.74 | 0.75 | 0.74 | 0.69 | 0.71 | 0.70 | 0.72 | 0.75 | 0.71 | 0.59 | 0.76 | 0.57 | 0.73 | 0.81 | 0.72 |
| Training on: UKA-CXR | 0.76 | 0.72 | 0.76 | 0.56 | 0.74 | 0.56 | 0.65 | 0.73 | 0.62 | 0.76 | 0.80 | 0.76 | 0.71 | 0.81 | 0.69 |
| Training on: PadChest | 0.77 | 0.74 | 0.78 | 0.67 | 0.70 | 0.68 | 0.68 | 0.72 | 0.66 | 0.66 | 0.66 | 0.66 | 0.77 | 0.79 | 0.76 |
| *Cross-institutional average* | *0.76* | *0.75* | *0.77* | *0.63* | *0.72* | *0.63* | *0.67* | *0.72* | *0.64* | *0.62* | *0.72* | *0.60* | *0.72* | *0.81* | *0.70* |
| $\varepsilon=\infty$ (Non-DP) | | | | | | | | | | | | | | | |
| Training on: VinDr-CXR | 0.85 | 0.86 | 0.85 | 0.58 | 0.73 | 0.58 | 0.67 | 0.70 | 0.65 | 0.63 | 0.69 | 0.62 | 0.70 | 0.81 | 0.68 |
| Training on: ChestX-ray14 | 0.76 | 0.74 | 0.75 | 0.71 | 0.72 | 0.72 | 0.71 | 0.73 | 0.70 | 0.69 | 0.61 | 0.70 | 0.76 | 0.80 | 0.75 |
| Training on: CheXpert | 0.81 | 0.75 | 0.80 | 0.68 | 0.72 | 0.70 | 0.74 | 0.77 | 0.73 | 0.60 | 0.76 | 0.58 | 0.76 | 0.79 | 0.75 |
| Training on: UKA-CXR | 0.80 | 0.78 | 0.77 | 0.62 | 0.69 | 0.62 | 0.68 | 0.70 | 0.66 | 0.79 | 0.82 | 0.78 | 0.71 | 0.81 | 0.69 |
| Training on: PadChest | 0.80 | 0.79 | 0.80 | 0.70 | 0.72 | 0.71 | 0.72 | 0.70 | 0.71 | 0.58 | 0.76 | 0.55 | 0.81 | 0.83 | 0.80 |
| *Cross-institutional average* | *0.79* | *0.77* | *0.78* | *0.65* | *0.72* | *0.65* | *0.70* | *0.71* | *0.68* | *0.63* | *0.71* | *0.61* | *0.73* | *0.80* | *0.72* |
| P-value | 0.21 | 0.51 | 0.48 | 0.38 | 0.89 | 0.29 | 0.005 | 0.10 | 0.009 | 0.82 | 0.76 | 0.84 | 0.31 | 0.39 | 0.30 |

**Table S5: Comparison of DP-DT and non-DP-DT in terms of fairness for different genders.** Values represent the demographic parity difference of the selected subgroup. Columns show test datasets, whereas rows represent training datasets. Diagnostic accuracy was calculated as an average over all labels including cardiomegaly, pleural effusion, pneumonia, atelectasis, and healthy. Sample sizes for different subgroups are reported in **Figures S1–S6**. Accuracy values for different subgroups are reported in **Tables S3** and **S4**. The privacy budgets of the DP networks corresponding to each dataset are as follows: VinDr-CXR: $\varepsilon=1.17$, ChestX-ray14: $\varepsilon=1.01$, CheXpert: $\varepsilon=0.98$, UKA-CXR: $\varepsilon=0.98$, and PadChest: $\varepsilon=0.72$, with $\delta = 0.000006$ for all datasets. P-values are calculated between DP and non-DP methods (for cross-institutional only). F=female, M=male.

|  | Test on: VinDr-CXR | | Test on: ChestX-ray14 | | Test on: CheXpert | | Test on: UKA-CXR | | Test on: PadChest | |
|---|---|---|---|---|---|---|---|---|---|---|
|  | F | M | F | M | F | M | F | M | F | M |
| $\varepsilon\approx1$ (DP) | | | | | | | | | | |
| Training on: VinDr-CXR | 0.01 | -0.01 | -0.02 | 0.02 | 0.01 | -0.01 | 0.00 | 0.00 | 0.05 | -0.05 |
| Training on: ChestX-ray14 | 0.01 | -0.01 | -0.02 | 0.02 | -0.01 | 0.01 | 0.04 | -0.04 | 0.04 | -0.04 |
| Training on: CheXpert | 0.02 | -0.02 | -0.01 | 0.01 | 0.01 | -0.01 | 0.04 | -0.04 | 0.03 | -0.03 |
| Training on: UKA-CXR | 0.02 | -0.02 | 0.00 | 0.00 | 0.00 | 0.00 | -0.01 | 0.01 | 0.04 | -0.04 |
| Training on: PadChest | 0.04 | -0.04 | 0.01 | -0.01 | -0.01 | 0.01 | -0.01 | 0.01 | 0.01 | -0.01 |
| *Cross-institutional average* | *0.02* | *-0.02* | *-0.01* | *0.01* | *0.00* | *0.00* | *0.02* | *-0.02* | *0.04* | *-0.04* |
| $\varepsilon=\infty$ (Non-DP) | | | | | | | | | | |
| Training on: VinDr-CXR | 0.05 | -0.05 | 0.02 | -0.02 | -0.02 | 0.02 | 0.01 | -0.01 | 0.05 | -0.05 |
| Training on: ChestX-ray14 | -0.01 | 0.01 | 0.00 | 0.00 | -0.01 | 0.01 | 0.02 | -0.02 | 0.01 | -0.01 |
| Training on: CheXpert | -0.06 | 0.06 | -0.01 | 0.01 | -0.01 | 0.0.1 | 0.01 | -0.01 | 0.01 | -0.01 |
| Training on: UKA-CXR | 0.02 | -0.02 | 0.00 | 0.00 | -0.04 | 0.04 | 0.00 | 0.00 | 0.02 | -0.02 |
| Training on: PadChest | 0.01 | -0.01 | -0.02 | 0.02 | -0.03 | 0.03 | 0.02 | -0.02 | 0.01 | -0.01 |
| *Cross-institutional average* | *-0.01* | *0.01* | *-0.01* | *0.01* | *-0.04* | *0.04* | *0.02* | *-0.02* | *0.02* | *-0.02* |
| P-value | 0.15 | 0.15 | 0.87 | 0.87 | 0.08 | 0.08 | 0.87 | 0.87 | 0.07 | 0.07 |

**Table S6: More detailed comparison of DP-DT and non-DP-DT in terms of cross-institutional performance for the subgroup with ages in the range of [0, 40) years.** Here we present the accuracy (ACC), sensitivity (Sens), and specificity (Spec) values for each training-testing combination, averaged over all one-vs-all classifications (cardiomegaly, pleural effusion, pneumonia, atelectasis, and healthy). Note that there is no significant difference between the DP-trained and the non-DP-trained networks. The privacy budgets of the DP networks are as follows: VinDr-CXR: $\varepsilon=1.17$, ChestX-ray14: $\varepsilon=1.01$, CheXpert: $\varepsilon=0.98$, UKA-CXR: $\varepsilon=0.98$, and PadChest: $\varepsilon=0.72$, with $\delta = 0.000006$ for all datasets. P-values were calculated between DP and non-DP methods (for cross-institutional only).

| [0, 40) years | Test on: VinDr-CXR | | | Test on: ChestX-ray14 | | | Test on: CheXpert | | | Test on: UKA-CXR | | | Test on: PadChest | | |
|---|---|---|---|---|---|---|---|---|---|---|---|---|---|---|---|
| | ACC | Sens | Spec | ACC | Sens | Spec | ACC | Sens | Spec | ACC | Sens | Spec | ACC | Sens | Spec |
| $\varepsilon \approx 1$ (DP) | | | | | | | | | | | | | | | |
| Training on: VinDr-CXR | 0.83 | 0.85 | 0.81 | 0.59 | 0.73 | 0.59 | 0.64 | 0.79 | 0.62 | 0.63 | 0.75 | 0.62 | 0.76 | 0.79 | 0.73 |
| Training on: ChestX-ray14 | 0.85 | 0.81 | 0.83 | 0.67 | 0.71 | 0.68 | 0.70 | 0.79 | 0.68 | 0.65 | 0.71 | 0.65 | 0.77 | 0.81 | 0.74 |
| Training on: CheXpert | 0.79 | 0.79 | 0.80 | 0.69 | 0.69 | 0.71 | 0.75 | 0.78 | 0.74 | 0.61 | 0.77 | 0.60 | 0.77 | 0.79 | 0.75 |
| Training on: UKA-CXR | 0.81 | 0.79 | 0.82 | 0.56 | 0.74 | 0.56 | 0.63 | 0.80 | 0.60 | 0.75 | 0.78 | 0.75 | 0.74 | 0.82 | 0.71 |
| Training on: PadChest | 0.83 | 0.80 | 0.84 | 0.67 | 0.69 | 0.69 | 0.67 | 0.79 | 0.65 | 0.67 | 0.72 | 0.67 | 0.75 | 0.82 | 0.73 |
| *Cross-institutional average* | *0.82* | *0.80* | *0.82* | *0.63* | *0.71* | *0.64* | *0.66* | *0.79* | *0.64* | *0.64* | *0.74* | *0.64* | *0.76* | *0.80* | *0.73* |
| $\varepsilon = \infty$ (Non-DP) | | | | | | | | | | | | | | | |
| Training on: VinDr-CXR | 0.88 | 0.88 | 0.87 | 0.59 | 0.73 | 0.59 | 0.63 | 0.79 | 0.60 | 0.68 | 0.67 | 0.69 | 0.73 | 0.80 | 0.70 |
| Training on: ChestX-ray14 | 0.79 | 0.79 | 0.79 | 0.72 | 0.72 | 0.74 | 0.72 | 0.77 | 0.71 | 0.62 | 0.75 | 0.61 | 0.77 | 0.78 | 0.74 |
| Training on: CheXpert | 0.81 | 0.79 | 0.81 | 0.68 | 0.73 | 0.69 | 0.77 | 0.79 | 0.76 | 0.67 | 0.75 | 0.66 | 0.77 | 0.80 | 0.75 |
| Training on: UKA-CXR | 0.82 | 0.80 | 0.79 | 0.62 | 0.71 | 0.62 | 0.65 | 0.77 | 0.62 | 0.81 | 0.85 | 0.80 | 0.74 | 0.78 | 0.72 |
| Training on: PadChest | 0.84 | 0.85 | 0.84 | 0.69 | 0.71 | 0.71 | 0.74 | 0.76 | 0.72 | 0.65 | 0.76 | 0.64 | 0.82 | 0.83 | 0.79 |
| *Cross-institutional average* | *0.82* | *0.81* | *0.81* | *0.65* | *0.72* | *0.65* | *0.69* | *0.77* | *0.66* | *0.66* | *0.73* | *0.65* | *0.75* | *0.79* | *0.73* |
| P-value | 0.80 | 0.55 | 0.30 | 0.34 | 0.65 | 0.44 | 0.23 | 0.07 | 0.27 | 0.56 | 0.87 | 0.64 | 0.39 | 0.41 | 0.60 |

**Table S7: More detailed comparison of DP-DT and non-DP-DT in terms of cross-institutional performance for the subgroup with ages in the range of [40, 70) years.** Here we present the accuracy (ACC), sensitivity (Sens), and specificity (Spec) values for each training-testing combination, averaged over all one-vs-all classifications (cardiomegaly, pleural effusion, pneumonia, atelectasis, and healthy). Note that there is no significant difference between the DP-trained and the non-DP-trained networks. The privacy budgets of the DP networks are as follows: VinDr-CXR: $\varepsilon=1.17$, ChestX-ray14: $\varepsilon=1.01$, CheXpert: $\varepsilon=0.98$, UKA-CXR: $\varepsilon=0.98$, and PadChest: $\varepsilon=0.72$, with $\delta = 0.000006$ for all datasets. P-values were calculated between DP and non-DP methods (for cross-institutional only).

| [40, 70) years | Test on: VinDr-CXR | | | Test on: ChestX-ray14 | | | Test on: CheXpert | | | Test on: UKA-CXR | | | Test on: PadChest | | |
|---|---|---|---|---|---|---|---|---|---|---|---|---|---|---|---|
| | ACC | Sens | Spec | ACC | Sens | Spec | ACC | Sens | Spec | ACC | Sens | Spec | ACC | Sens | Spec |
| $\varepsilon \approx 1$ (DP) | | | | | | | | | | | | | | | |
| Training on: VinDr-CXR | 0.77 | 0.76 | 0.77 | 0.59 | 0.70 | 0.59 | 0.64 | 0.72 | 0.61 | 0.65 | 0.73 | 0.62 | 0.74 | 0.77 | 0.73 |
| Training on: ChestX-ray14 | 0.78 | 0.76 | 0.78 | 0.68 | 0.68 | 0.69 | 0.67 | 0.74 | 0.66 | 0.60 | 0.74 | 0.57 | 0.75 | 0.81 | 0.73 |
| Training on: CheXpert | 0.79 | 0.70 | 0.78 | 0.67 | 0.70 | 0.69 | 0.72 | 0.75 | 0.71 | 0.61 | 0.75 | 0.59 | 0.77 | 0.79 | 0.76 |
| Training on: UKA-CXR | 0.79 | 0.75 | 0.78 | 0.57 | 0.71 | 0.57 | 0.65 | 0.72 | 0.63 | 0.76 | 0.79 | 0.75 | 0.73 | 0.78 | 0.72 |
| Training on: PadChest | 0.78 | 0.72 | 0.79 | 0.66 | 0.69 | 0.68 | 0.67 | 0.73 | 0.66 | 0.64 | 0.70 | 0.63 | 0.77 | 0.80 | 0.77 |
| *Cross-institutional average* | *0.79* | *0.73* | *0.78* | *0.62* | *0.70* | *0.63* | *0.66* | *0.73* | *0.64* | *0.63* | *0.73* | *0.60* | *0.75* | *0.79* | *0.74* |
| $\varepsilon = \infty$ (Non-DP) | | | | | | | | | | | | | | | |
| Training on: VinDr-CXR | 0.85 | 0.81 | 0.85 | 0.59 | 0.71 | 0.59 | 0.67 | 0.69 | 0.65 | 0.64 | 0.70 | 0.63 | 0.75 | 0.77 | 0.74 |
| Training on: ChestX-ray14 | 0.80 | 0.74 | 0.81 | 0.71 | 0.71 | 0.72 | 0.70 | 0.75 | 0.69 | 0.72 | 0.58 | 0.74 | 0.78 | 0.80 | 0.76 |
| Training on: CheXpert | 0.78 | 0.73 | 0.80 | 0.69 | 0.70 | 0.71 | 0.73 | 0.78 | 0.72 | 0.61 | 0.76 | 0.59 | 0.78 | 0.79 | 0.76 |
| Training on: UKA-CXR | 0.81 | 0.70 | 0.79 | 0.62 | 0.67 | 0.63 | 0.66 | 0.71 | 0.64 | 0.80 | 0.83 | 0.79 | 0.74 | 0.76 | 0.72 |
| Training on: PadChest | 0.80 | 0.77 | 0.79 | 0.69 | 0.72 | 0.70 | 0.71 | 0.72 | 0.70 | 0.59 | 0.77 | 0.56 | 0.83 | 0.83 | 0.82 |
| *Cross-institutional average* | *0.80* | *0.74* | *0.80* | *0.65* | *0.70* | *0.66* | *0.69* | *0.72* | *0.67* | *0.64* | *0.70* | *0.63* | *0.76* | *0.78* | *0.75* |
| P-value | 0.19 | 0.92 | 0.10 | 0.10 | 0.99 | 0.14 | 0.02 | 0.31 | 0.02 | 0.71 | 0.61 | 0.62 | 0.06 | 0.21 | 0.25 |

**Table S8: More detailed comparison of DP-DT and non-DP-DT in terms of cross-institutional performance for the subgroup with ages in the range of [70, 100) years.** Here we present the accuracy (ACC), sensitivity (Sens), and specificity (Spec) values for each training-testing combination, averaged over all one-vs-all classifications (cardiomegaly, pleural effusion, pneumonia, atelectasis, and healthy). Note that there is no significant difference between the DP-trained and the non-DP-trained networks. The privacy budgets of the DP networks are as follows: VinDr-CXR: $\varepsilon=1.17$, ChestX-ray14: $\varepsilon=1.01$, CheXpert: $\varepsilon=0.98$, UKA-CXR: $\varepsilon=0.98$, and PadChest: $\varepsilon=0.72$, with $\delta = 0.000006$ for all datasets. P-values were calculated between DP and non-DP methods (for cross-institutional only).

| [70, 100) years | Test on: VinDr-CXR | | | Test on: ChestX-ray14 | | | Test on: CheXpert | | | Test on: UKA-CXR | | | Test on: PadChest | | |
|---|---|---|---|---|---|---|---|---|---|---|---|---|---|---|---|
| | ACC | Sens | Spec | ACC | Sens | Spec | ACC | Sens | Spec | ACC | Sens | Spec | ACC | Sens | Spec |
| $\varepsilon \approx 1$ (DP) | | | | | | | | | | | | | | | |
| Training on: VinDr-CXR | 0.78 | 0.69 | 0.78 | 0.60 | 0.65 | 0.61 | 0.66 | 0.66 | 0.64 | 0.63 | 0.68 | 0.62 | 0.67 | 0.77 | 0.65 |
| Training on: ChestX-ray14 | 0.76 | 0.68 | 0.75 | 0.67 | 0.65 | 0.69 | 0.69 | 0.68 | 0.68 | 0.63 | 0.65 | 0.62 | 0.70 | 0.78 | 0.69 |
| Training on: CheXpert | 0.73 | 0.69 | 0.73 | 0.68 | 0.67 | 0.70 | 0.71 | 0.72 | 0.70 | 0.67 | 0.62 | 0.68 | 0.70 | 0.77 | 0.69 |
| Training on: UKA-CXR | 0.77 | 0.66 | 0.78 | 0.55 | 0.70 | 0.53 | 0.60 | 0.74 | 0.57 | 0.74 | 0.76 | 0.74 | 0.65 | 0.81 | 0.63 |
| Training on: PadChest | 0.76 | 0.65 | 0.76 | 0.68 | 0.63 | 0.71 | 0.68 | 0.69 | 0.66 | 0.65 | 0.64 | 0.65 | 0.74 | 0.75 | 0.74 |
| *Cross-institutional average* | *0.76* | *0.67* | *0.76* | *0.63* | *0.66* | *0.64* | *0.66* | *0.69* | *0.64* | *0.65* | *0.65* | *0.64* | *0.68* | *0.78* | *0.67* |
| $\varepsilon = \infty$ (Non-DP) | | | | | | | | | | | | | | | |
| Training on: VinDr-CXR | 0.83 | 0.73 | 0.83 | 0.58 | 0.66 | 0.58 | 0.64 | 0.67 | 0.62 | 0.62 | 0.66 | 0.61 | 0.68 | 0.75 | 0.67 |
| Training on: ChestX-ray14 | 0.77 | 0.75 | 0.77 | 0.69 | 0.68 | 0.70 | 0.68 | 0.71 | 0.67 | 0.64 | 0.65 | 0.64 | 0.72 | 0.76 | 0.72 |
| Training on: CheXpert | 0.78 | 0.68 | 0.80 | 0.70 | 0.67 | 0.71 | 0.73 | 0.74 | 0.72 | 0.63 | 0.70 | 0.62 | 0.72 | 0.77 | 0.71 |
| Training on: UKA-CXR | 0.75 | 0.64 | 0.75 | 0.57 | 0.69 | 0.57 | 0.65 | 0.69 | 0.63 | 0.78 | 0.79 | 0.78 | 0.66 | 0.79 | 0.65 |
| Training on: PadChest | 0.76 | 0.73 | 0.76 | 0.66 | 0.69 | 0.67 | 0.70 | 0.67 | 0.69 | 0.57 | 0.75 | 0.54 | 0.76 | 0.82 | 0.75 |
| *Cross-institutional average* | *0.77* | *0.70* | *0.77* | *0.63* | *0.68* | *0.63* | *0.67* | *0.69* | *0.65* | *0.62* | *0.69* | *0.60* | *0.70* | *0.77* | *0.69* |
| P-value | 0.55 | 0.33 | 0.53 | 0.99 | 0.41 | 0.80 | 0.57 | 0.70 | 0.48 | 0.22 | 0.27 | 0.26 | 0.01 | 0.06 | 0.003 |

**Table S9: Comparison of DP-DT and non-DP-DT in terms of fairness for each age subgroup.** Values represent the demographic parity difference of the selected subgroup. Columns show test datasets, whereas rows represent training datasets. Diagnostic accuracy was calculated as an average over all labels including cardiomegaly, pleural effusion, pneumonia, atelectasis, and healthy. Sample sizes for different subgroups are reported in **Figures S1–S6**. Accuracy values for different age subgroups are reported in **Tables S6–S8**. P-values are calculated between DP and non-DP methods (for cross-institutional only). Age subgroups are given in years.

| | Test on: VinDr-CXR | | | Test on: ChestX-ray14 | | | Test on: CheXpert | | | Test on: UKA-CXR | | | Test on: PadChest | | |
|---|---|---|---|---|---|---|---|---|---|---|---|---|---|---|---|
| | [0, 40) | [40, 70) | [70, 100) | [0, 40) | [40, 70) | [70, 100) | [0, 40) | [40, 70) | [70, 100) | [0, 40) | [40, 70) | [70, 100) | [0, 40) | [40, 70) | [70, 100) |
| $\varepsilon \approx 1$ (DP) | | | | | | | | | | | | | | | |
| Training on: VinDr-CXR | 0.06 | -0.04 | -0.01 | 0.00 | 0.00 | 0.01 | -0.01 | -0.01 | 0.02 | -0.01 | 0.02 | -0.02 | 0.05 | 0.04 | -0.08 |
| Training on: ChestX-ray14 | 0.07 | -0.04 | -0.05 | -0.01 | 0.01 | -0.01 | 0.02 | -0.02 | 0.01 | 0.04 | -0.03 | 0.02 | 0.04 | 0.03 | -0.06 |
| Training on: CheXpert | 0.01 | 0.02 | -0.06 | 0.02 | -0.02 | 0.00 | 0.03 | 0.00 | -0.02 | -0.03 | -0.05 | 0.06 | 0.03 | 0.05 | -0.07 |
| Training on: UKA-CXR | 0.02 | -0.01 | -0.03 | -0.01 | 0.01 | -0.02 | 0.00 | 0.04 | 0.05 | 0.00 | 0.02 | -0.02 | 0.04 | 0.05 | -0.08 |
| Training on: PadChest | 0.05 | -0.03 | -0.04 | 0.01 | -0.01 | 0.02 | 0.00 | -0.01 | 0.01 | 0.03 | -0.01 | 0.01 | -0.01 | 0.03 | -0.02 |
| *Cross-institutional average* | *0.04* | *-0.02* | *-0.05* | *0.01* | *-0.01* | *0.00* | *0.00* | *0.00* | *0.02* | *0.01* | *-0.02* | *0.02* | *0.04* | *0.04* | *-0.07* |
| $\varepsilon = \infty$ (Non-DP) | | | | | | | | | | | | | | | |
| Training on: VinDr-CXR | 0.03 | -0.01 | -0.03 | 0.00 | 0.00 | -0.01 | -0.03 | 0.03 | -0.02 | 0.05 | 0.01 | -0.02 | 0.01 | 0.05 | -0.06 |
| Training on: ChestX-ray14 | 0.00 | 0.02 | -0.03 | 0.01 | 0.00 | -0.01 | 0.03 | 0.01 | -0.02 | -0.06 | 0.08 | -0.07 | 0.01 | 0.04 | -0.06 |
| Training on: CheXpert | 0.03 | -0.02 | -0.01 | -0.01 | 0.01 | 0.01 | 0.04 | -0.01 | -0.01 | 0.05 | -0.02 | 0.01 | 0.01 | 0.04 | -0.06 |
| Training on: UKA-CXR | 0.02 | 0.01 | -0.06 | 0.00 | 0.01 | -0.05 | -0.01 | 0.01 | -0.01 | 0.02 | 0.02 | -0.02 | 0.03 | 0.05 | -0.08 |
| Training on: PadChest | 0.05 | -0.01 | -0.05 | 0.00 | 0.01 | -0.03 | 0.03 | 0.00 | -0.02 | 0.07 | 0.01 | -0.03 | 0.02 | 0.05 | -0.07 |
| *Cross-institutional average* | *0.03* | *0.00* | *-0.04* | *0.00* | *0.01* | *-0.02* | *0.01* | *0.01* | *-0.02* | *0.03* | *0.02* | *-0.03* | *0.02* | *0.05* | *-0.07* |
| P-value | 0.57 | 0.52 | 0.70 | 0.44 | 0.19 | 0.17 | 0.84 | 0.48 | 0.01 | 0.66 | 0.24 | 0.09 | 0.03 | 0.64 | 0.21 |

**Ablation Study**

To enhance the applicability of our results, we expanded our experiments to include two more network architectures: ResNet18[26], encompassing 11,179,077 parameters, and EfficientNet B0[27], containing 4,013,953 parameters.

Except for the batch normalization layers, which we replaced with group normalization, our method closely followed the original designs of ResNet18 by He et al. [26], and EfficientNet B0 by Tan et al[27]. Consistently, the training parameters employed for ResNet9 were replicated for both differential privacy (DP) and non-DP training for these architectures. This includes parameters such as δ, batch size, the maximum gradient norm, DP accountant, learning rates, optimizers, data augmentation strategies, as well as training and test data dimensions.

A comprehensive evaluation of the results, depicted in terms of average area under the receiver operating characteristic curve (AUC) across labels such as cardiomegaly, pleural effusion, pneumonia, atelectasis, and healthy classifications, can be found in **Figures S10** and **S11** for ResNet18 and **Figures S12** and **S13** for EfficientNet B0. For both differential privacy-enhanced domain transfer (DP-DT) with $\varepsilon \approx 1$ (VinDr-CXR: $\varepsilon = 1.15$, ChestX-ray14: $\varepsilon = 0.70$, CheXpert: $\varepsilon = 0.98$, UKA-CXR: $\varepsilon = 0.88$, and PadChest: $\varepsilon = 0.75$ for ResNet18 and VinDr-CXR: $\varepsilon = 1.03$, ChestX-ray14: $\varepsilon = 0.58$, CheXpert: $\varepsilon = 0.58$, UKA-CXR: $\varepsilon = 0.78$, and PadChest: $\varepsilon = 0.58$ for EfficientNet B0) and $\delta = 6 \times 10^{-6}$ and non-DP-DT ($\varepsilon = \infty$) across all datasets and diverse demographic subsets, the trends observed echo those witnessed with ResNet9.

We initially chose cardiomegaly, pleural effusion, pneumonia, atelectasis, and the healthy category as the target labels because these were the only ones consistently available across all datasets utilized in our research. To extend the scope of our results, we looked for additional imaging findings prevalent in most datasets. As such, pneumothorax and consolidation emerged as suitable choices, given their presence in the VinDr-CXR, ChestX-ray14, CheXpert, and PadChest datasets. We conducted further experiments using the ResNet9 architecture, maintaining the same configuration and training parameters as in our primary setup. The AUC results for pneumothorax and consolidation for both DP-DT with $\varepsilon \approx 1$ (VinDr-CXR: $\varepsilon = 1.07$, ChestX-ray14: $\varepsilon = 0.87$, CheXpert: $\varepsilon = 0.96$, and PadChest: $\varepsilon = 0.98$) and $\delta = 6 \times 10^{-6}$ and non-DP-DT ($\varepsilon = \infty$) are displayed in **Figure S14**. Consistent with the primary findings of our study, these results exhibited a similar trend.

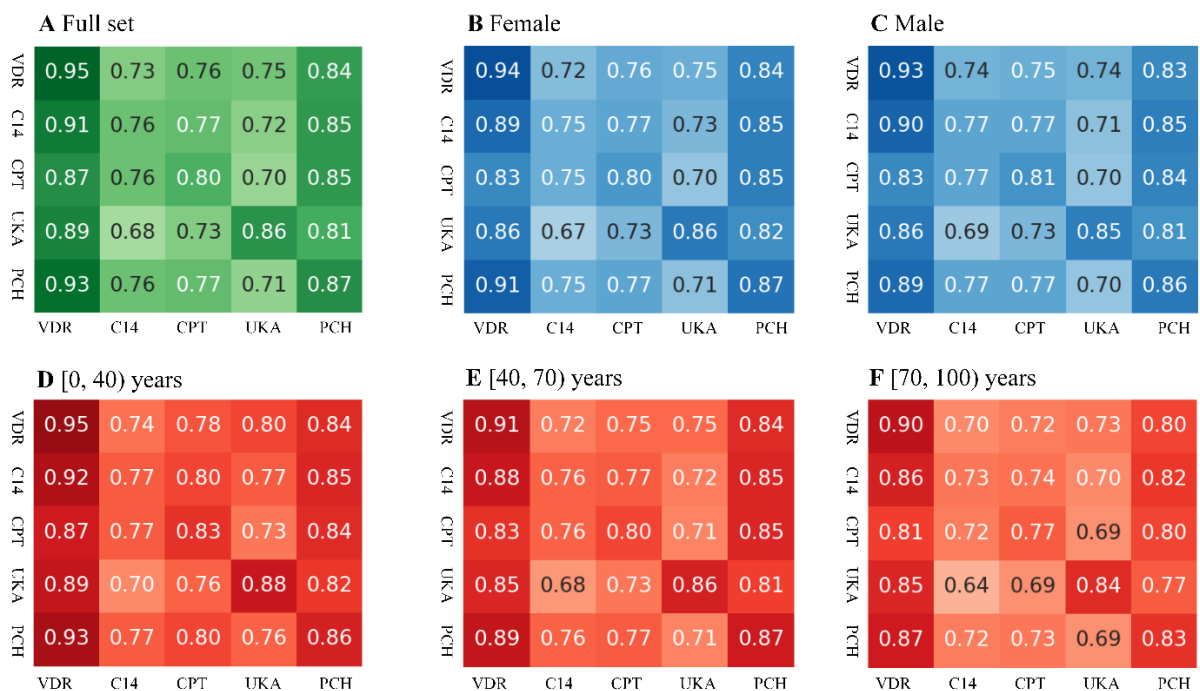

**Figure S10**: **Results of transferring differential privacy (DP) models with ε≈1 to different domains in terms of average area under the receiver operating characteristic curve (AUC) using ResNet18 architecture.** Each row corresponds to a training domain and each column corresponds to a test domain. The results represent average AUC values over all labels including cardiomegaly, pleural effusion, pneumonia, atelectasis, and healthy participants. The AUC values correspond to **(A)** full test sets, **(B)** female subgroups, **(C)** male subgroups, **(D)** [0, 40) years subgroups, **(E)** [40, 70) years subgroups, and **(F)** [70, 100) years subgroups. The privacy budgets of the DP networks corresponding to each dataset are as follows: VinDr-CXR (VDR): $\varepsilon = 1.15$, ChestX-ray14 (C14): $\varepsilon = 0.70$, CheXpert (CPT): $\varepsilon = 0.98$, UKA-CXR (UKA): $\varepsilon = 0.88$, and PadChest (PCH): $\varepsilon = 0.75$ with $\delta = 0.000006$ for all datasets. DP-DT=differential privacy-enhanced domain transfer.

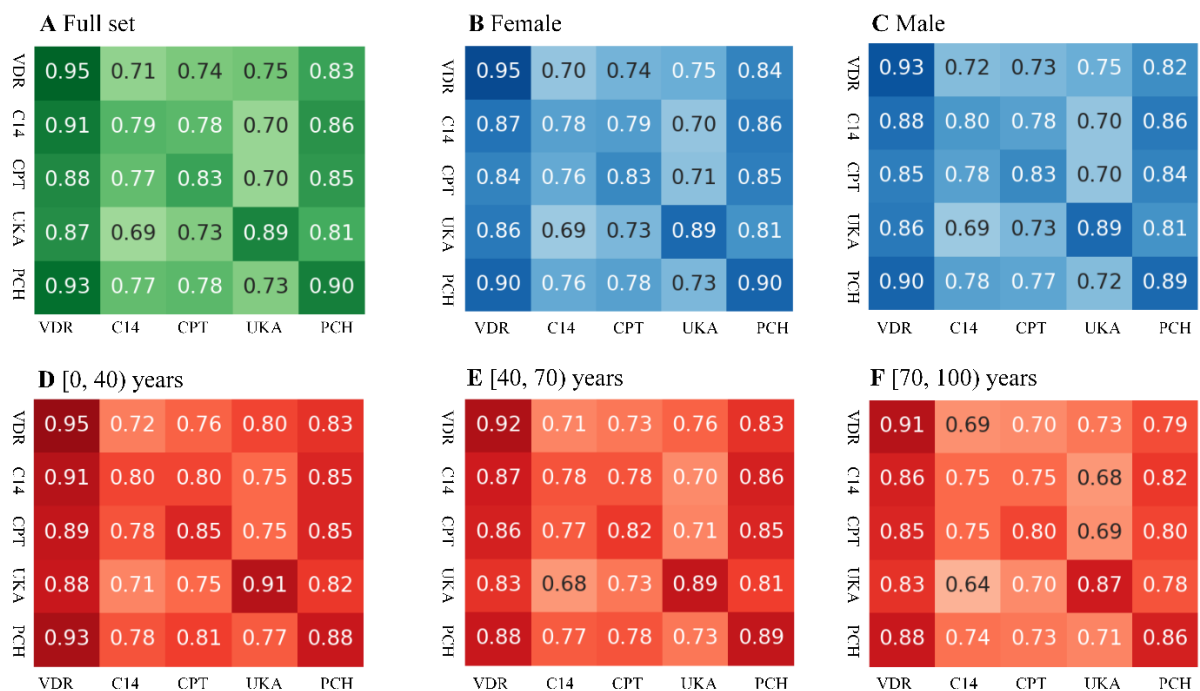

Figure S11: **Results of transferring non-DP models (ε=∞) to different domains in terms of average area under the receiver operating characteristic curve (AUC) using ResNet18 architecture.** Each row corresponds to a training domain and each column corresponds to a test domain. The results represent average AUC values over all labels including cardiomegaly, pleural effusion, pneumonia, atelectasis, and healthy participants. The AUC values correspond to **(A)** full test sets, **(B)** female subgroups, **(C)** male subgroups, **(D)** [0, 40) years subgroups, **(E)** [40, 70) years subgroups, and **(F)** [70, 100) years subgroups. DP-DT=differential privacy-enhanced domain transfer, VDR=VinDr-CXR, C14=ChestX-ray14, CPT=CheXpert, UKA=UKA-CXR, PCH=PadChest.

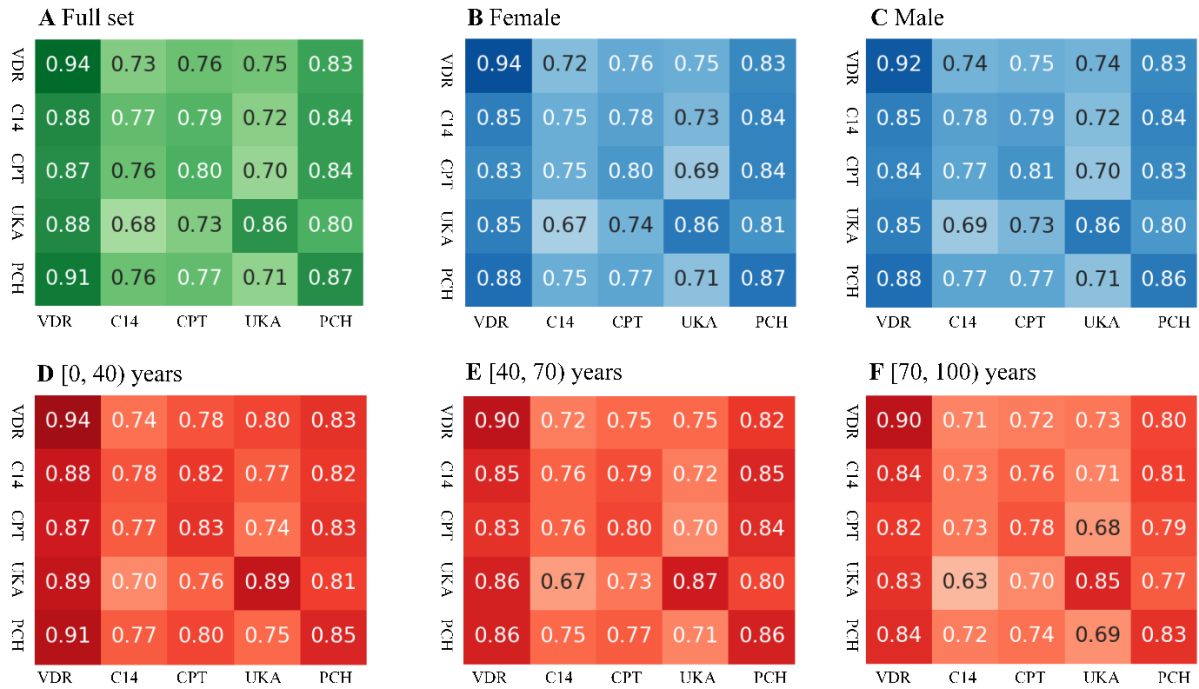

**Figure S12**: **Results of transferring DP models with ε≈1 to different domains in terms of average area under the receiver operating characteristic curve (AUC) using EfficientNet B0 architecture.** Each row corresponds to a training domain and each column corresponds to a test domain. The results represent average AUC values over all labels including cardiomegaly, pleural effusion, pneumonia, atelectasis, and healthy participants. The AUC values correspond to **(A)** full test sets, **(B)** female subgroups, **(C)** male subgroups, **(D)** [0, 40) years subgroups, **(E)** [40, 70) years subgroups, and **(F)** [70, 100) years subgroups. The privacy budgets of the DP networks corresponding to each dataset are as follows: VinDr-CXR (VDR): $\varepsilon = 1.03$, ChestX-ray14 (C14): $\varepsilon = 0.58$, CheXpert (CPT): $\varepsilon = 0.58$, UKA-CXR (UKA): $\varepsilon = 0.78$, and PadChest (PCH): $\varepsilon = 0.58$ with $\delta = 0.000006$ for all datasets. DP-DT=differential privacy-enhanced domain transfer.

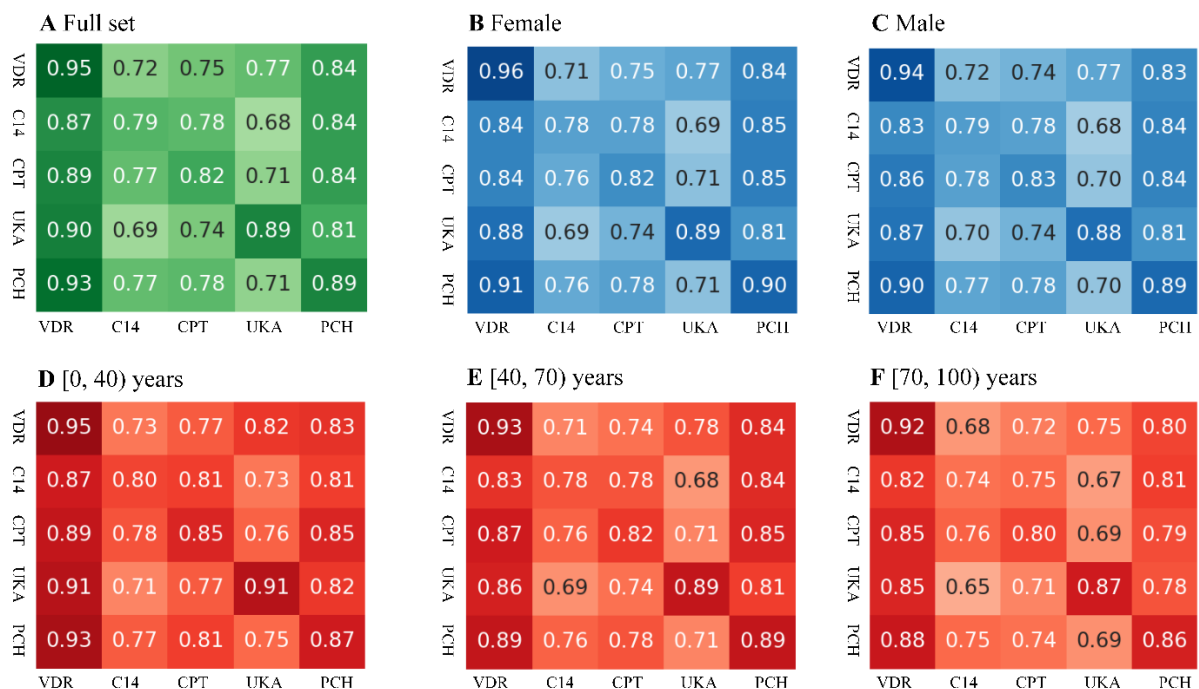

Figure S13: **Results of transferring non-DP models (ε=∞) to different domains in terms of average area under the receiver operating characteristic curve (AUC) using EfficientNet B0 architecture.** Each row corresponds to a training domain and each column corresponds to a test domain. The results represent average AUC values over all labels including cardiomegaly, pleural effusion, pneumonia, atelectasis, and healthy participants. The AUC values correspond to **(A)** full test sets, **(B)** female subgroups, **(C)** male subgroups, **(D)** [0, 40) years subgroups, **(E)** [40, 70) years subgroups, and **(F)** [70, 100) years subgroups. DP-DT=differential privacy-enhanced domain transfer, VDR=VinDr-CXR, C14=ChestX-ray14, CPT=CheXpert, UKA=UKA-CXR, PCH=PadChest.

## AUC values for DP-DT (ε ≈ 1) and non-DP-DT (ε = ∞)
## For Pneumothorax and Consolidation using ResNet9

**A** Pneumothorax (ε ≈ 1)

|  | VDR | C14 | CPT | PCH |
|---|---|---|---|---|
| VDR | 0.93 | 0.70 | 0.74 | 0.78 |
| C14 | 0.92 | 0.78 | 0.76 | 0.85 |
| CPT | 0.86 | 0.83 | 0.83 | 0.87 |
| PCH | 0.88 | 0.76 | 0.77 | 0.87 |

**B** Consolidation (ε ≈ 1)

|  | VDR | C14 | CPT | PCH |
|---|---|---|---|---|
| VDR | 0.90 | 0.68 | 0.70 | 0.84 |
| C14 | 0.89 | 0.69 | 0.70 | 0.86 |
| CPT | 0.91 | 0.70 | 0.73 | 0.88 |
| PCH | 0.92 | 0.70 | 0.72 | 0.88 |

**C** Pneumothorax (ε = ∞)

|  | VDR | C14 | CPT | PCH |
|---|---|---|---|---|
| VDR | 0.95 | 0.74 | 0.75 | 0.83 |
| C14 | 0.84 | 0.86 | 0.79 | 0.81 |
| CPT | 0.89 | 0.85 | 0.87 | 0.83 |
| PCH | 0.92 | 0.75 | 0.79 | 0.92 |

**D** Consolidation (ε = ∞)

|  | VDR | C14 | CPT | PCH |
|---|---|---|---|---|
| VDR | 0.92 | 0.70 | 0.71 | 0.85 |
| C14 | 0.89 | 0.74 | 0.70 | 0.85 |
| CPT | 0.91 | 0.70 | 0.74 | 0.87 |
| PCH | 0.92 | 0.70 | 0.71 | 0.90 |

**Figure S14**: **Results of transferring DP models (ε≈1) and non-DP models (ε=∞) to different domains in terms of AUC for pneumothorax and consolidation using ResNet9 architecture.** Each row corresponds to a training domain and each column corresponds to a test domain. The area under the receiver operating characteristic curve (AUC) values correspond to **(A)** pneumothorax according to DP-DT with $\varepsilon \approx 1$, **(B)** consolidation according to DP-DT with $\varepsilon \approx 1$, **(C)** pneumothorax according to non-DP-DT ($\varepsilon = \infty$), and **(D)** consolidation according to non-DP-DT ($\varepsilon = \infty$). The privacy budgets of the DP networks corresponding to each dataset are as follows: VinDr-CXR (VDR): ε=1.07, ChestX-ray14 (C14): ε=0.87, CheXpert (CPT): ε=0.96, and PadChest (PCH): ε=0.98, with $\delta = 0.000006$ for all datasets. DP-DT=differential privacy-enhanced domain transfer.

## Computational Efficiency Analysis

In this section, we evaluated the computational time needed for training with and without differential privacy (DP) across different network architectures and datasets. While the computational demands during network testing remained consistent between DP and non-DP-trained models, training times varied. **Table S10** detailed the training convergence time, in terms of total training time, for various DP and non-DP scenarios, all on identical hardware. On average, training with DP took 3.6 times longer for ResNet9[26,30], 5.3 times longer for ResNet18[26], and 11.7 times longer for EfficientNet B0[27].

**Table S10**: **Comparison of computational training times across datasets with and without differential privacy (DP).** Values represent convergence times in minutes. DP networks operated with $\varepsilon \approx 1$, with $\delta = 0.000006$ for all datasets. Results span three distinct network architectures including ResNet9, ResNet18, and EfficientNet B0, and the ratio of DP to non-DP training times is also provided.

| Dataset | ResNet9 | | | ResNet18 | | | EfficientNet B0 | | |
|---|---|---|---|---|---|---|---|---|---|
| | DP [minutes] | Non-DP [minutes] | DP / Non-DP | DP [minutes] | Non-DP [minutes] | DP / Non-DP | DP [minutes] | Non-DP [minutes] | DP / Non-DP |
| VinDr-CXR | 34 | 22 | 1.5 | 65 | 13 | 5.0 | 136 | 10 | 13.6 |
| ChestX-ray14 | 491 | 66 | 7.4 | 346 | 40 | 8.6 | 768 | 83 | 9.3 |
| CheXpert | 1148 | 206 | 5.6 | 1185 | 180 | 6.6 | 1137 | 63 | 18.0 |
| UKA-CXR | 772 | 279 | 2.8 | 996 | 267 | 3.7 | 1027 | 196 | 5.2 |
| PadChest | 210 | 236 | 0.9 | 399 | 141 | 2.8 | 792 | 63 | 12.6 |
| *Average* | *531* | *162* | *3.6* | *598* | *128* | *5.3* | *772* | *83* | *11.7* |